\documentclass[letterpaper, 10 pt, conference]{ieeeconf}  

\IEEEoverridecommandlockouts                              

\usepackage{tabu}
\usepackage{graphicx}
\usepackage{newclude}
\usepackage{color}
\usepackage{xcolor}
\usepackage[colorlinks,linkcolor=blue]{hyperref}
\usepackage{etoolbox}
\usepackage{subfigure}
\usepackage{amsmath}
\usepackage{ulem}
\usepackage{tikz}
\usepackage{cite}
\usepackage{url}
\usepackage{verbatim}
\usepackage{booktabs}
\usepackage{pifont}
\usepackage{algorithm}
\usepackage{algpseudocode}
\usetikzlibrary{positioning}
\usetikzlibrary{matrix}
\usepackage{pgfplots}
\usepackage{pgfplotstable}
\usepackage{tikz}
\tikzset{>=stealth}
\usetikzlibrary{patterns}
\usetikzlibrary{pgfplots.statistics}
\usetikzlibrary{positioning, shapes}
\usetikzlibrary{calc}
\usetikzlibrary{backgrounds,scopes}
\usetikzlibrary{fit}
\usetikzlibrary{arrows.meta}
\makeatletter
\patchcmd{\@makecaption}
  {\scshape}
  {}
  {}
  {}
\makeatletter
\patchcmd{\@makecaption}
  {\\}
  {.\ }
  {}
  {}
\makeatother

\title{\LARGE \bf
Learning from Symmetry: Meta-Reinforcement Learning with Symmetrical Behaviors and Language Instructions
}

\author{Xiangtong~Yao$^{1}$,
	Zhenshan~Bing$^{1}$,
	Genghang~Zhuang$^{1}$,
	Kejia~Chen$^{1}$,
	Hongkuan~Zhou$^{1}$,\\
	Kai~Huang$^{2}$ $^{3}$ and
	Alois~Knoll$^{1}$
	\thanks{$^{1}$ School of Computation, Information and Technology, Technical University of Munich, Germany.
		E-mail: yaox@in.tum.de}
	\thanks{$^{2}$ School of Computer Science and Engineering, Sun Yat-sen University, China. Email: huangk36@mail.sysu.edu.cn}
	\thanks{{$^{3}$} Pazhou Lab, Guangzhou, 510330, China}
	}
\providecommand{\keywords}[1]{\textbf{\textit{Keywords---}} #1}
\begin{document}

\maketitle
\thispagestyle{empty}
\pagestyle{empty}

\newcommand{\hwplotA}{\raisebox{2pt}{\tikz{\draw[black,solid,line width=1.2pt](0,0) -- (5mm,0);}}}
\newcommand{\hwplotB}{\raisebox{2pt}{\tikz{\draw[black,dashed,line width=1.2pt](0,0) -- (5mm,0);}}}
\newcommand{\hwplotC}{\raisebox{2pt}{\tikz{\draw[blue,solid,line width=1.2pt](0,0) -- (5mm,0);}}}
\newcommand{\hwplotD}{\raisebox{2pt}{\tikz{\draw[magenta,dashed,line width=1.2pt](0,0) -- (5mm,0);}}}

\newcommand{\hwplotE}{\raisebox{2pt}{\tikz{\draw[->, yellow,solid,line width=1.2pt](0,0) -- (5mm,0);}}}
\newcommand{\hwplotF}{\raisebox{2pt}{\tikz{\fill[red](0,0) circle(0.8mm);}}}
\newcommand{\hwplotG}{\raisebox{2pt}{\tikz{\fill[green](0,0) circle(0.8mm);}}}
\newcommand{\hwplotH}{\raisebox{2pt}{\tikz{\draw[green,dashed,line width=1.2pt](0,0) circle(0.8mm);}}}
\newcommand{\hwplotI}{\raisebox{2pt}{\tikz{\draw[red,dash dot,line width=1.2pt](0,0) -- (5mm,0);}}}
\renewcommand{\algorithmicrequire}{\textbf{Input:}}  
\renewcommand{\algorithmicensure}{\textbf{Output:}} 

\tikzstyle{arrow} = [thick,->,>=stealth,rounded corners=4pt, draw=black, align=center]
\tikzstyle{arrow1} = [line width=0.2mm,->,>=stealth,rounded corners=4pt, draw=black!70, align=center]
\tikzstyle{origin} = [circle,fill=black,thick,inner sep=0pt, minimum size=1mm]
\tikzstyle{target1} = [circle,fill=green,thick,inner sep=0pt, minimum size=1mm]
\tikzstyle{target2} = [circle,fill=orange,thick,inner sep=0pt, minimum size=1mm]
\tikzstyle{target3} = [circle,fill=red,thick,inner sep=0pt, minimum size=1mm]
\tikzstyle{endpoint} = [circle,fill=black,thick,inner sep=0pt, minimum size=0.5mm]
\tikzstyle{arcstyle} = [start angle=0, end angle=180, radius=10mm]
\tikzstyle{arrow2} = [thick,<->,>=stealth,rounded corners=4pt, draw=black, align=center]

\begin{abstract}
Meta-reinforcement learning (meta-RL) is a promising approach that enables the agent to learn new tasks quickly. However, most meta-RL algorithms show poor generalization in multi-task scenarios due to the insufficient task information provided only by rewards. Language-conditioned meta-RL improves the generalization capability by matching language instructions with the agent's behaviors. While both behaviors and language instructions have symmetry, which can speed up human learning of new knowledge. Thus, combining symmetry and language instructions into meta-RL can help improve the algorithm's generalization and learning efficiency. We propose a dual-MDP meta-reinforcement learning method that enables learning new tasks efficiently with symmetrical behaviors and language instructions. We evaluate our method in multiple challenging manipulation tasks, and experimental results show that our method can greatly improve the generalization and learning efficiency of meta-reinforcement learning. Videos are available at \href{https://tumi6robot.wixsite.com/symmetry/}{{https://tumi6robot.wixsite.com/symmetry/}}.
\end{abstract}

\section{Introduction}

\textcolor{black}{Meta-reinforcement learning (meta-RL) is a learning approach that mimics the human learning process to empower agents with the ability to understand and adapt to new knowledge. During meta-training, the agent engages in intensive interactions with Markov decision process (MDP) environments, learning tasks, and generalizing to new and unseen tasks from the same task family\cite{finn2017model,rakelly2019efficient,wang2023metareinforcement,bing2023diva,9804728}. However, most meta-RL agents face challenges in adapting to diverse sets of tasks, such as Meta-World\cite{yu2020meta}, as they solely rely on rewards to make decisions and adapt to new tasks. These rewards, however, are not sufficiently informative for the agents to effectively transfer prior knowledge to new tasks.}

The application of natural language to meta-RL has resulted in the development of language-conditioned meta-RL. This approach resembles the process of teaching by iteratively describing tasks and providing corrective feedback to the agent, enabling the agent to connect new tasks with prior knowledge and demonstrates good generalization in challenging task sets and complex scenarios\cite{co2018guiding}. However, language-conditioned meta-RL is often time-consuming and inefficient due to the requirement of matching language instructions and behaviors through trial-and-error process.

\textcolor{black}{In order to enhance the generalization and efficiency of meta-RL algorithms, other forms of human learning are being explored. One such approach involves considering symmetry as a carrier of information. Studies have shown that humans, including infants, can distinguish symmetrical visual patterns\cite{symmetryinfant, PORNSTEIN19851}, suggesting that learning from symmetry plays a key role in human learning. Therefore, incorporating symmetry into meta-RL has been put forth as a means to accelerate the learning process and improve generalization of new tasks\cite{zhou2020meta, kirsch2022introducing}. However, these approaches are applicable to simple tasks, such as the bandit problem, not to complex manipulation tasks. This is due to the complexity of agent behaviors required to solve multi-goal manipulation tasks \cite{9772990, 9466373}. These complex behaviors present substantial modeling challenges when trying to adopt symmetry-based approaches.}

Combining symmetry and language instructions in meta-RL has the potential to improve learning efficiency and generalization in complex manipulation tasks, much like the human learning process. For instance, during the process of learning how to open a drawer with guidance from a teacher, one may naturally attempt to close the drawer as well without explicit instructions, as the actions of opening and closing are symmetrical. People also have the ability to spontaneously explore symmetry, making it easier to learn the closing action from the opening action. If the teacher informs the learner that the closing action corresponds to the task of closing the drawer, they will quickly learn how to perform this task.

However, few studies have focused on the combination of symmetry and language instructions in robotic manipulation tasks at present. The reasons are twofold. Firstly, incorporating symmetry or language instructions into meta-RL is a relatively new area of research and lacks a general framework to address the problem. Secondly, most existing methods learn the symmetry knowledge from meta-training tasks, and then use this knowledge to directly reshape the reward function of new tasks. This approach works in simple scenarios where the reward functions of new tasks and meta-training tasks are similar. However, in robotic manipulation tasks, the reward functions of new tasks and meta-training tasks are often significantly different\cite{yu2020meta}, making it difficult to reshape the reward function of new tasks directly through the symmetry knowledge of meta-training tasks.

This paper presents a novel investigation into combining symmetry and language instructions in meta-RL for the manipulation tasks problem. In particular, the proposed method focuses on integrating symmetrical language instructions with the symmetrical behaviors of meta-training tasks to learn meta-test tasks in robotic manipulation tasks. The key contributions of this work are summarized as follows:
\begin{itemize}
    \item \textcolor{black}{A symmetry-based method has been developed for multiple manipulation tasks. This method generates symmetrical language instructions and symmetrical behaviors, which assists the agent in disambiguating the keywords in language instructions and accurately matching language instructions with the appropriate behaviors.}
    \item \textcolor{black}{A dual-MDP meta-RL scheme based on language-conditioned meta-RL is proposed, which incorporates the symmetry-based method to accelerate the meta-test adaptation of the agent. The experimental results indicate that our approach can effectively improve both the generalization and efficiency of learning new manipulation tasks. In particular, our method achieves a 43.7\% higher success rate compared to MILLION\cite{Million} in meta-test tasks within the multi-manipulation task scenario.}
\end{itemize}

\section{Related Work}

In recent years, language-conditioned (LC) model-free methods have been successful in learning multiple robotic manipulation tasks, including LC reinforcement learning\cite{goyal2021pixl2r}, LC imitation learning \cite{jang2022bc, shridhar2022cliport,zhou2023languageconditioned} and LC meta-RL\cite{co2018guiding, Million}. These approaches consider the unstructured language instructions as a part of task formulation. To be specific, the task objective is conveyed through language instructions and these instructions are encoded into the observation space, which enables the agent to interact with the environment in accordance with the intended purpose of the instructions, resulting in an accelerated learning process and well generalization for multiple manipulation tasks.

\textcolor{black}{Despite the aforementioned successes of LC model-free methods, the reliance on extensive expert demonstrations\cite{jang2022bc, shridhar2022cliport} or carefully crafted reward functions\cite{Million} as a means of grounding language instructions with manipulation tasks remains a prominent limitation. In addition, while these approaches may enable the agent to solve new tasks that were not encountered during the training phase, the success rate of these solutions is relatively low. A primary reason is that the keywords of language instructions are highly abstract and context-specific in nature, often leading to differing interpretations of the same keywords in different scenarios. For example, the instruction ``\textbf{open} the door" will elicit a distinct behavior from that of the instruction ``\textbf{open} the drawer". The failure of the agent to solve the latter task through the transfer of concepts learned from the former task highlights the difficulty in accurately grounding language instructions with the correct behavior. While the exploration of the behavioral characteristics of training tasks, such as symmetry, will aid the agent in correctly grounding the language instructions with the behavior for new tasks.}

Zinkevich et al. theoretically define symmetry in MDPs and propose that symmetry is conducive to accelerating learning \cite{zinkevich2001symmetry}. Subsequently, symmetry is introduced into RL or meta-RL, utilizing the symmetry in various ways \cite{kirsch2022introducing, agostini2009exploiting, van2020mdp, lin2020invariant}. In the field of robotic manipulation tasks, a natural approach to explore symmetry is augmenting the training data with behavioral symmetry, e.g., Lin et al. utilized symmetrical reflection to augment manipulation task trajectories, which proved effective in accelerating agent exploration and exploitation, achieving promising results for tasks such as pushing, sliding, and pick-and-place~\cite{lin2020invariant}. \textcolor{black}{However, its restricted domain, with the training and test tasks belonging to the same task set, renders it inapplicable to meta-RL problems. Specifically, it cannot solve a novel task, such as closing a door, through symmetrical reflection of the behavior learned from an opening door task, if the closing door task is not appeared in the training phase. To extend the applicability of symmetry to language-conditioned meta-RL, we present a novel training scheme aimed at improving the association between language instructions and appropriate behaviors in the multi-manipulation task scenario.}

\section{Background}
\textcolor{black}{This work proposes a dual-MDP method for learning new tasks from symmetrical data generated by meta-training tasks based on language-conditioned meta-RL.}

Language-conditioned Meta-RL aims to train an agent, with the input of language instructions, on a meta-training set $D^{train}_T$ to quickly adapt to new unseen tasks from a meta-test set $D^{test}_T$, where the meta-training and the meta-test tasks are drawn from the same task distribution $p(T)$~\cite{bing2022TPAMI}. MILLION\cite{Million} is a language-conditioned meta-RL that incorporates free-form language instructions and trial-and-error mechanism to improve the algorithm's generalization, showing start-of-the-art performance on Meta-World benchmark. Thus, MILLION forms the baseline of our proposed method.

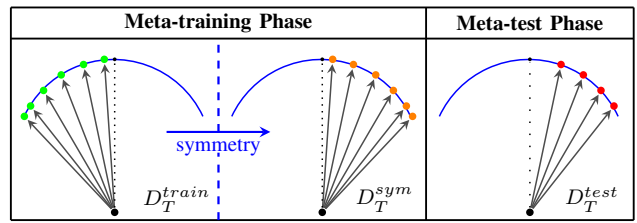
\begin{figure}[t!]
  \centering
    \begin{tikzpicture}[scale=0.69]
        \draw [line width=0.25mm] (0, 0) -- (12, 0);
        \draw [line width=0.25mm] (0, 3.5) -- (12, 3.5);
        \draw [line width=0.25mm] (0, 0) -- (0, 4.1);
        \draw [line width=0.3mm, dashed, blue] (4, 0) -- (4, 1.3);
        \draw [line width=0.3mm, dashed, blue] (4, 1.7) -- (4, 3.5);
        \draw [line width=0.25mm] (8, 0) -- (8, 4.1);
        \draw [line width=0.25mm] (12, 0) -- (12, 4.1);
        \draw [line width=0.25mm] (0, 4.1) -- (12, 4.1);

        \draw[blue, line width=0.2mm] (3.7,2) arc (25:155:1.9);
        \draw[blue, line width=0.2mm] (7.7,2) arc (25:155:1.9);
        \draw[blue, line width=0.2mm] (11.7,2) arc (25:155:1.9);

        \node(end1) at (2,3.1) [endpoint] {};
        \node(end2) at (6,3.1) [endpoint] {};
        \node(end3) at (10,3.1) [endpoint] {};
        \node(train) at (2,0.15) [origin] {};
        \node(sym) at (6,0.15) [origin] {};
        \node(test) at (10,0.15) [origin] {};
        \node at (3.2,0.4) {\footnotesize $D^{train}_T$};
        \node at (7.2,0.4) {\footnotesize $D^{sym}_T$};
        \node at (11.2,0.4) {\footnotesize $D^{test}_T$};

        \node(train10) at (0.26,2)[target1]{};
        \node(train11) at (0.38,2.2)[target1]{};
        \node(train12) at (0.63,2.5)[target1]{};
        \node(train13) at (0.97,2.8)[target1]{};
        \node(train14) at (1.4,3)[target1]{};
        \node(train15) at (1.8,3.1)[target1]{};

        \draw[arrow1] (train) -- (train10);
        \draw[arrow1] (train) -- (train11);
        \draw[arrow1] (train) -- (train12);
        \draw[arrow1] (train) -- (train13);
        \draw[arrow1] (train) -- (train14);
        \draw[arrow1] (train) -- (train15);

        \node(train20) at (7.74,2)[target2]{};
        \node(train21) at (7.62,2.2)[target2]{};
        \node(train22) at (7.37,2.5)[target2]{};
        \node(train23) at (7.03,2.8)[target2]{};
        \node(train24) at (6.6,3)[target2]{};
        \node(train25) at (6.2,3.1)[target2]{};

        \draw[arrow1] (sym) -- (train20);
        \draw[arrow1] (sym) -- (train21);
        \draw[arrow1] (sym) -- (train22);
        \draw[arrow1] (sym) -- (train23);
        \draw[arrow1] (sym) -- (train24);
        \draw[arrow1] (sym) -- (train25);

        \node(train31) at (11.62,2.2)[target3]{};
        \node(train32) at (11.37,2.5)[target3]{};
        \node(train33) at (11.03,2.8)[target3]{};
        \node(train34) at (10.6,3)[target3]{};

        \draw[arrow1] (test) -- (train31);
        \draw[arrow1] (test) -- (train32);
        \draw[arrow1] (test) -- (train33);
        \draw[arrow1] (test) -- (train34);

        \draw[arrow,blue] (3,1.7) -- (5,1.7);
        \node at (4,1.4) [blue] {\footnotesize symmetry};

        \draw[black,line width=0.2mm, dotted] (train) -- (end1);
        \draw[black,line width=0.2mm, dotted] (sym) -- (end2);
        \draw[black,line width=0.2mm,loosely dotted] (test) -- (end3);

        \node at (4,3.8) {\textbf{\footnotesize Meta-training Phase}};
        \node at (10,3.8) {\textbf{\footnotesize Meta-test Phase}};
    \end{tikzpicture}
  \caption{The symmetrical task set $D^{sym}_T$ is augmented by $D^{train}_T$. The tasks in $D^{sym}_T$ bear resemblance to those in $D^{test}_T$, yet they are not identical. Different color dots on the arc represent different tasks.}\label{fig.sym}
  \label{fig.training}
  \vspace{-10pt}
\end{figure}

\begin{figure*}[!htbp]
  \centering
  \includegraphics[width=0.95\linewidth]{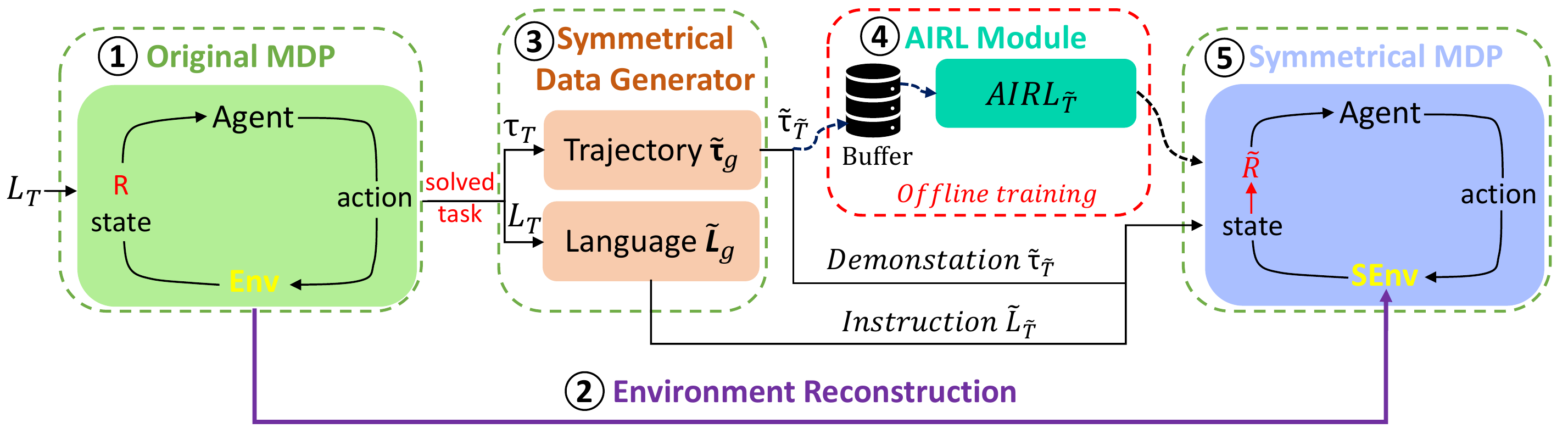}
  \caption{The scheme of dual-MDP meta-RL. We use MILLION\cite{Million} to train the agent. Specifically, The training process in \textbf{Original MDP} is consistent with that of MILLION. $R$ is the reward function. $T$ and $\tilde{T}$ represent meta-training tasks and symmetrical tasks respectively. $L$ indicates language instructions, and $\tilde{L}$ represents symmetrical language instructions. $\tau$ and $\tilde{\tau}$ represent the trajectory of meta-training tasks and symmetrical tasks respectively. \textbf{Buffer} gathers the symmetrical trajectory $\tilde{\tau}$. $\tilde{\tau}_g$ is the symmetrical trajectory generator, $\tilde{L}_g$ is the symmetrical language instructions generator.}
  \label{fig.framework}
\end{figure*}

\section{Methodology}

Our goal is to train an agent on a meta-training set $D^{train}_T$ and a symmetric task set $D^{sym}_T$, such that the agent can efficiently learn new tasks of $D^{test}_T$, where $D^{sym}_T$ is augmented from $D^{train}_T$ with the behavioral symmetry of task $T \in D^{train}_T$, and the tasks of $D^{sym}_T$ resemble those of $D^{test}_T$, shown as Figure~\ref{fig.training}. Consequently, the agent performs the meta-training tasks' MDPs and the symmetrical tasks' MDPs in the meta-training stage simultaneously. Then the agent adapts quickly to the new tasks of $D^{test}_T$ in the meta-test stage. The scheme of our method is shown in Figure~\ref{fig.framework}.

The architecture consists of five modules, including an original MDP module, a symmetrical data generator, an AIRL module, an environment reconstructing module and a symmetrical MDP module. The process is as follows:
\begin{enumerate}
\item[\ding{192}] The original MDP module receives language instruction $L$ and carries out meta-training tasks' MDPs to solve task and generate the successful trajectory $\tau$.
\item[\ding{193}] The environment reconstruction module builds the symmetrical task environment $SEnv$ by reconstructing the meta-training task environment $Env$.
\item[\ding{194}] \textcolor{black}{The symmetrical data generator receives $\tau$ and $L$ of the meta-training tasks and generates the symmetrical trajectory $\tilde{\tau}$ and the symmetrical language instruction $\tilde{L}$ by utilizing the symmetrical trajectory generator $\tilde{\tau}_g$ and the symmetrical language instructions generator $\tilde{L}_g$, respectively.}
\item[\ding{195}] \textcolor{black}{In the AIRL module, the replay buffer, namely \textbf{Buffer}, is employed to gather the symmetrical trajectory $\tilde{\tau}$, which will be utilized by offline training AIRL to recover the symmetrical task's reward function $\tilde{R}$.}
\item[\ding{196}] \textcolor{black}{The symmetrical MDP module receives $\tilde{L}$ and $\tilde{\tau}$, and employs $\tilde{R}$ and $SEnv$ to perform symmetrical task MDPs. In addition, $\tilde{\tau}$ serves as a demonstration, facilitating the agent in efficiently solving the symmetrical task $\tilde{T}$ $\in$ $D^{sym}_T$. Finally, the agent adapts to meta-test tasks in $D^{test}_T$ in the meta-test phase.}
\end{enumerate}
\textcolor{black}{The overview algorithm of our method is shown in \textbf{Algorithm}~\ref{alg::dualmdp} and \textbf{Algorithm}~\ref{alg::dualmdp2}. In this section, we provide a formal definition of symmetrical tasks and provide an in-depth explanation of each module.}

\begin{figure*}[htbp!]
  \centering
  \sbox0{\hwplotE}\sbox1{\hwplotF}\sbox2{\hwplotG}\sbox3{\hwplotH}

  \begin{minipage}[b]{0.16\linewidth}
    \centerline{\includegraphics[width=\linewidth,height=1.3\linewidth]{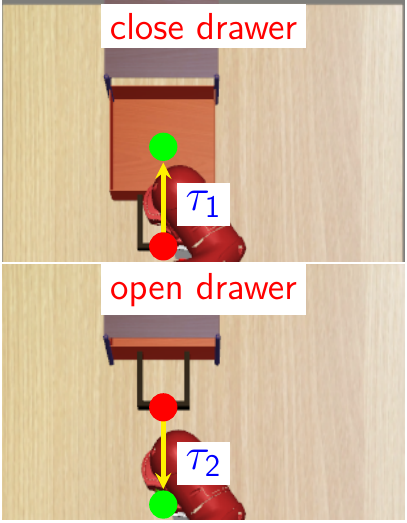}}
    \centerline{\quad \textcolor{black}{\footnotesize (a) Drawer task}}
  \end{minipage}
  \begin{minipage}[b]{0.16\linewidth}
    \centerline{\includegraphics[width=\linewidth,height=1.3\linewidth]{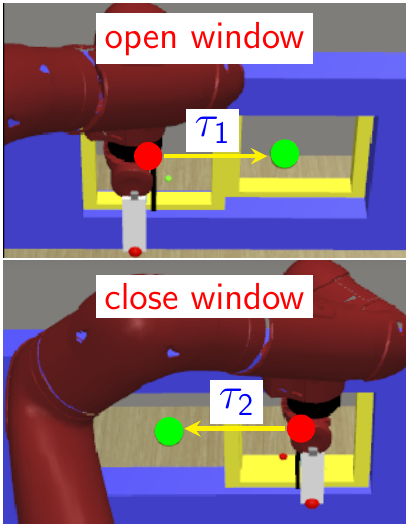}}
    \centerline{\quad \textcolor{black}{\footnotesize (b) Window task}}
  \end{minipage}
  \begin{minipage}[b]{0.16\linewidth}
    \centerline{\includegraphics[width=\linewidth,height=1.3\linewidth]{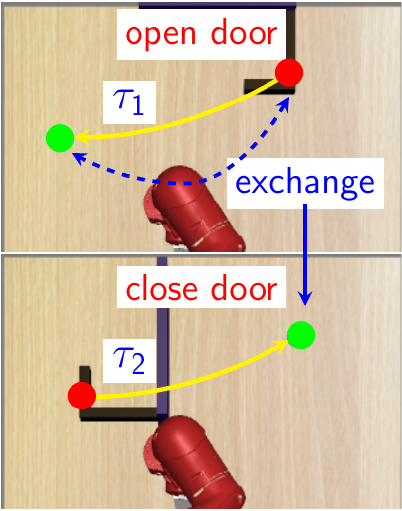}}
    \centerline{\quad \textcolor{black}{\footnotesize (c) Door task}}
  \end{minipage}
  \begin{minipage}[b]{0.16\linewidth}
    \centerline{\includegraphics[width=\linewidth,height=1.3\linewidth]{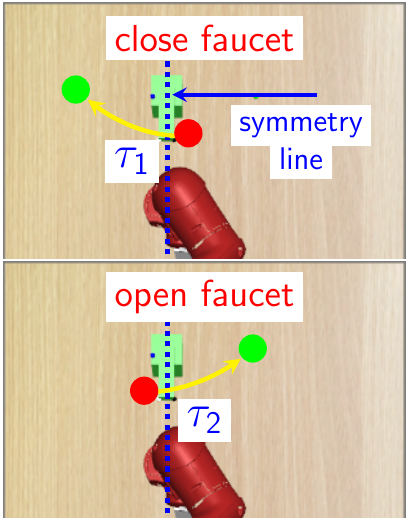}}
    \centerline{\quad \textcolor{black}{\footnotesize (d) Faucet task}}
  \end{minipage}
  \begin{minipage}[b]{0.16\linewidth}
    \centerline{\includegraphics[width=\linewidth,height=1.3\linewidth]{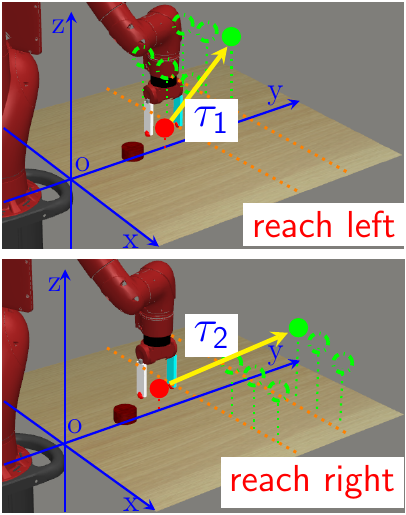}}
    \centerline{\quad \textcolor{black}{\footnotesize (e) Reaching task}}
  \end{minipage}
  \begin{minipage}[b]{0.16\linewidth}
    \centerline{\includegraphics[width=\linewidth,height=1.3\linewidth]{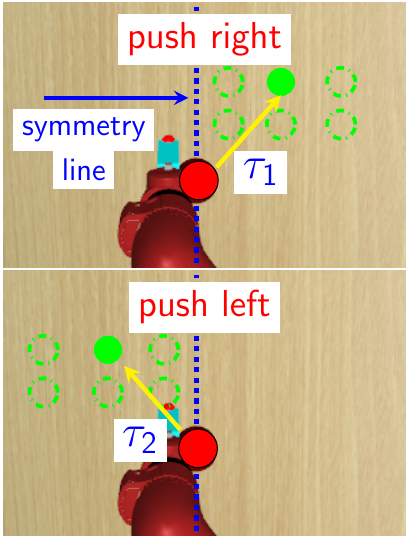}}
    \centerline{\quad \textcolor{black}{\footnotesize (f) Pushing task}}
  \end{minipage}

  \vspace{-5pt}
  \caption{Six symmetrical task families visualized in the Mujoco simulation environment\cite{todorov2012mujoco}, and these tasks are selected from Meta-World ML45 benchmark\cite{yu2020meta}. The \textbf{first row} are \textbf{meta-training tasks}, and the \textbf{second row} are the corresponding \textbf{meta-test tasks}. In each task family, the meta-training task and meta-test task are symmetrical to each other. \usebox1 represents the object key point, \usebox2 indicates the task goal. The yellow line shows the agent's path as it solves the task by controlling the key point to reach the task goal. In addition, \usebox3 in the \textbf{reach} and \textbf{push} tasks indicates multiple task goals.}\label{fig.task}
  \vspace{-10pt}
\end{figure*}

\begin{algorithm}[t!]
  \caption{\textcolor{black}{Dual-MDP Meta-RL: Meta-training}}
  \label{alg::dualmdp}
  \begin{algorithmic}[1]
    \Require $L$: Language instructions;
    \Require $D^{train}_T$: Meta-training tasks;

    \State Initial data buffer $D$;
    \Repeat
        \For{$T \in D^{train}_T$}
        \State Environment $Env$ of $T$;
        \Repeat
            \State Perform $\pi(a|s,L_T)$ on $T$ in OMDP;
        \Until{the agent solves the task $T$}
        \State Collect successful trajectory of $T$: $\tau \leftarrow$ OMDP;
        \State $SEnv \in \tilde{T} $ is reconstructed by $Env$;
        \State Generate $\tilde{\tau} \leftarrow \tilde{\tau}_g(\tau)$ and $\tilde{L} \leftarrow \tilde{L}_g(L)$;
        \State \textcolor{black}{Gather the symmetrical trajectory $\tilde{\tau}$ to \textbf{Buffer}};
        \EndFor
    \Until{The size of \textbf{Buffer} exceeds 4000}
    \State \textcolor{black}{Offline training AIRL to recover a reward function for $\tilde{T}$: $\tilde{R} \leftarrow AIRL(\textbf{Buffer}(\tilde{\tau}))$};

    \While{algorithm not converged}
        \For{$T \in D^{train}_T$}
            \State Perform $\pi(a|s,L_T)$ on $T$ in OMDP;
            \State Collect trajectory $\tau$ to $D$;
            \State Perform $\pi(a|s,\tilde{L}_{\tilde{T}})$ on $\tilde{T}$ in SMDP with $\tilde{R}$;
            \State Collect trajectory $\tilde{\tau}$ to $D$;
        \EndFor
        \State Train $\pi$ on $D$;
    \EndWhile
  \end{algorithmic}
\end{algorithm}

\begin{algorithm}[t!]
  \caption{\textcolor{black}{Dual-MDP Meta-RL: Meta-test}}
  \label{alg::dualmdp2}
  \begin{algorithmic}[1]
    \Require $L$: Language instructions;
    \Require $D^{test}_T$: Meta-test tasks;

    \For{$T\in D^{test}_T$}
        \State Perform $\pi(a|s,L_T)$ on $T$;
    \EndFor

  \end{algorithmic}
\end{algorithm}

\subsection{Symmetrical Tasks}
\noindent \textbf{Definition 1:} Assume that there are two tasks sampled from the same task family. The agent solves these two tasks in individual MDPs, resulting in two trajectories $\tau_1$ and $\tau_2$, respectively. As illustrated in Figure~\ref{fig.task}.(c) and Figure~\ref{fig.task}.(d), if, after an exchange or symmetry operation, the initial point and terminal point of $\tau_1$ coincide with those of $\tau_2$, then $\tau_1$ and $\tau_2$ are symmetrical to each other, and the tasks represented by these two trajectories are symmetrical tasks.

Figure~\ref{fig.task} visualizes 12 common symmetrical tasks in daily life. The yellow line with arrow represents the trajectory of the agent solving the task, and the red point symbolizes the object key point $P_{key}$ that the agent manipulates to solve the task. The green point indicates the task goal $P_{goal}$ that is to be achieved. $P_{key}$ and $P_{goal}$ serve as the initial and terminal points, respectively, as defined in \textbf{Definition 1}. The entire trajectory $\tau$ can be partitioned into two components: reaching trajectory $\tau_e$ and controlling trajectory $\tau_c$. $\tau_e$ represents the path of the agent from its initial state to reach the object key point $P_{key}$, while $\tau_c$ indicates the path of the agent in manipulating the object key point $P_{key}$ to attain the task goal $P_{goal}$. Specifically,
\vspace{-3pt}
\begin{flalign}
\label{equ:tau0}
\nonumber
\tau &= \tau_e + \tau_c&& \\ \nonumber
 &= ((s_0,a_0,r_0),(s_1,a_1,r_1),...,(s_e,a_e,r_e),...,s_N)&& \\
s &= [gp_x, gp_y, gp_z, obj_x, obj_y, obj_z] \\\nonumber
a &= [\Delta x, \Delta y, \Delta z, f] &&\nonumber
\end{flalign}
where action $a$ represents the displacement and the force of the gripper in a time step, state $s$ records the position of the gripper $gp$ and the object $obj$, $e$ is the time step at which the agent reaches the key point $P_{key}$, and $N$ is the maximum time step in the MDP's horizon.

\textcolor{black}{The two transformation rules of \textbf{Definition 1} correspond to daily life experience of individuals who  manipulate objects (e.g. drawer, door, window), where their behavior is constrained by the inherent movement rules of the object. It should be noted, however, that the symmetry rule defined in Definition 1 may not be generalizable to all manipulation tasks. In this work, we demonstrate the use of symmetry as a means of improving the grounding of language instructions and behaviors, utilizing the tasks depicted in Figure~\ref{fig.task} as a case study, rather than proposing a general approach to tackle all manipulation tasks. Our primary contribution lies in providing a new scheme that integrating symmetrical behaviors and symmetrical language instructions to accelerate and improve the agent's learning of new manipulation tasks.}

\subsection{Original MDP and Symmetrical MDP}

\textcolor{black}{In this work, it is assumed that the aforementioned symmetrical tasks are episodic task, rather than temporally-extended task, thus they are learned by the agent in separate, individual MDPs. This separation is necessary to ensure the accuracy of learning different tasks, as combining information from multiple tasks in a single MDP would lead to confusion for the agent. The original MDP module (OMDP) handles the meta-training task MDPs, while the symmetrical MDP module (SMDP) is responsible for setting up the symmetrical task MDPs, as depicted in Figure~\ref{fig.framework}. }

\subsection{Environment Reconstruction}
\textcolor{black}{The aim of reconstructing the environment $SEnv$ for the symmetrical task $\tilde{T}$ is to enable the agent to solve $\tilde{T}$ through interactions with $SEnv$. While the environment reconstruction is achieved by adjusting the state of the object and the position of the object key point $P_{key}$ and the task goal $P_{goal}$ in the original environment $Env$. If the agent successfully solves the meta-training task such that the final state of the object is similar to the initial state of the object in the corresponding meta-test task, such as Figure~\ref{fig.task}(a)-(c), then $SEnv$ can be reconstructed by preserving the object's final state and exchanging the $P_{key}$ and $P_{goal}$. Alternatively, if the initial state of the object is similar between the meta-training and meta-testing tasks, $SEnv$ can be reconstructed by maintaining the initial state of the object and making $P_{key}$ and $P_{goal}$ symmetrical about a symmetry line that is parallel to the $y$-axis and crosses through the object's midpoint, as shown in Figure~\ref{fig.task}(e)-(f).}

\textcolor{black}{In order maintain the generality of the meta-RL setting by ensuring that meta-test tasks do not appear during the meta-training phase, the task settings in Figure~\ref{fig.task} are consistent with those of the ML45 benchmark\cite{yu2020meta}. In particular, the height and position of the objects in the meta-training tasks differ from those in the corresponding meta-test tasks. As a result, the restructured  environment $SEnv$ is distinct from that of the meta-test task, thereby the agent's solution to the symmetrical task during the meta-training phase does not affect the integrity of the meta-RL setting.}

\subsection{Symmetrical Data Generator}

The symmetrical data generator module is composed of two sub-modules: symmetrical trajectory generator and symmetrical language instructions generator.

\subsubsection{Symmetrical trajectory generator}

\begin{figure}[htbp!]
  \centering
  \sbox0{\hwplotE}\sbox1{\hwplotF}\sbox2{\hwplotG}\sbox3{\hwplotH}

  \begin{minipage}[b]{0.4\linewidth}
    \centerline{\includegraphics[width=\linewidth,height=1.3\linewidth]{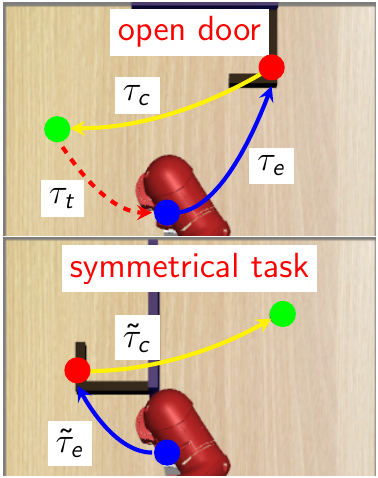}}
    \centerline{\quad \textcolor{black}{\footnotesize (a) Door task}}
  \end{minipage}
  \hspace{5pt}
  \begin{minipage}[b]{0.4\linewidth}
    \centerline{\includegraphics[width=\linewidth,height=1.3\linewidth]{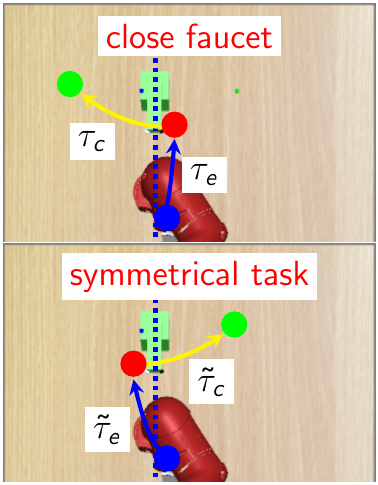}}
    \centerline{\quad \textcolor{black}{\footnotesize (b) Faucet task}}
  \end{minipage}

  \caption{Symmetrical trajectories generation. The blue point represents the initial position of the gripper.  $\tau_e$ and $\tau_c$ correspond to the trajectory of the gripper that the agent from its initial state to the successful completion of the task. $\tau_t$ indicates the trajectory of the gripper returning to the initial state after the meta-training task is completed.}\label{fig.sym-task}
  \vspace{-10pt}
\end{figure}

\textcolor{black}{The symmetrical trajectory generator converts the successful trajectory $\tau$ of the meta-training task to the symmetrical task trajectory $\tilde{\tau}$. Specifically, the converting process includes converting the reaching trajectory $\tau_e$ to $\tilde{\tau}_e$ and converting the controlling trajectory $\tau_c$ to $\tilde{\tau}_c$ in accordance with Equation~\ref{equ:tau0}, as shown in Figure~\ref{fig.sym-task}. If the symmetrical task environment, $SEnv$, is constructed by exchanging $P_{key}$ and $P_{obj}$ in the meta-training task environment $Env$, the trajectory of the agent in solving these two tasks is inverse to each other, i.e., $\tau_c$ and $\Tilde{\tau}_c$ in Figure~\ref{fig.sym-task}(a). For the sake of formal consistency, that is, the entire trajectory of the symmetrical task is inverse to that of the training task, we consider building an auxiliary trajectory of the training task, $\tau_t$, which indicates the trajectory of the gripper returning to the initial state after the training task is solved. Therefore, the entire symmetrical trajectory is:
\begin{flalign}
    \Tilde{\tau}&=\Tilde{\tau}_e+\Tilde{\tau}_c=Revise(\tau_c+\tau_t)&&\\ \nonumber
    &= ((s_t,a^*_t,r^*_t),(s_{t-1},a^*_{t-1},r^*_{t-1}),...,(s_N,a^*_N,r^*_N),...,s_{e})&& \\
    a^*_t &= -a_t = [-\Delta x_t, -\Delta y_t, \Delta z_t, f_t] && \nonumber
    \label{equ:tau1}
\end{flalign}
where $a^*_t$ is an action vector at step $t$, and its $\Delta x$ and $\Delta y$ are opposite to those of $a_t$, $r^*_t$ is obtained from the recovered reward function $\tilde{R}$ in the AIRL module.}

\textcolor{black}{While if $SEnv$ is obtained by the symmetry of $P_{key}$ and $P_{obj}$ with respect of the symmetry line, then the symmetrical trajectory can be generated by symmetry operation about the same symmetry line, as shown in Figure~\ref{fig.sym-task}(b). The symmetrical trajectory can be expressed as:
\begin{flalign}
    \Tilde{\tau}&=\Tilde{\tau}_e+\Tilde{\tau}_c=Symmetry(\tau_c+\tau_t)&&\\ \nonumber
    &= ((s_0,a^{+}_0,r^{+}_0),(s^{+}_1,a^{+}_1,r^{+}_1),...,(s^{+}_e,a^{+}_e,r^{+}_e),...,s^{+}_N)&& \\
    a^{+}_0 &= -a_0 = [-\Delta x_0, \Delta y_0, \Delta z_0, f_0] && \nonumber
    \label{equ:tau2}
\end{flalign}
The $y$ coordinates of the two trajectories are in one-to-one corresponding at the each time step as the symmetry line is parallel to the $y$-axis. Consequently, the item $\Delta x$ in $a^+$ is negative counterpart of that in $a$. Additionally, the symmetrical task and the corresponding meta-training task share the same dynamic function $P\{s_{t+1}|s_t,a_t\}$ in MDP. The symmetrical action $a^+_t$ results in the symmetrical state $s^+_t$, which is reflection of the original state $s_t$ with respect to the symmetry line. $r^{+}$ is obtained from $\tilde{R}$.}

\subsubsection{Symmetrical language instructions generator}
\textcolor{black}{This generator interprets the language instruction $L$ of the meta-training task $T$ to generate the symmetrical language instruction $\tilde{L}$, which is of an opposite meaning to $L$. In this work, the tasks are artificially described by language instructions. As these tasks are commonplace in daily life, the language instructions used for description are not complicated. Further details are available on our website\protect\footnotemark[1].}

\begin{figure}[htbp!]
  \centering
    \begin{tikzpicture}[level distance=1.3cm,
        level 1/.style={sibling distance=4cm, level distance=1cm},
        level 2/.style={sibling distance=1.5cm, level distance=0.8cm}]
        \node {Push forward the drawer handle}
        child {node {VP}
        child {node {Push}}
        child {node {Forward}}
        }
        child {node {NP}
        child {node {The}}
        child {node {Drawer}}
        child {node {Handle}}
        };
    \end{tikzpicture}
  \caption{Constituency parse for the language instruction ``push forward the drawer handle''. VP represents Verb Phrase and NP represents Noun Phrase.}\label{fig.parse}
\end{figure}
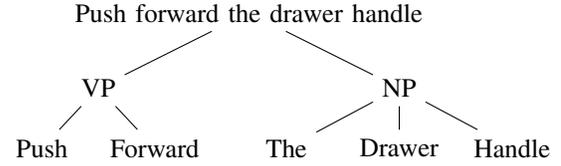

\textcolor{black}{As an illustration, consider the drawer-close task in Figure~\ref{fig.task}(a), where $L$ is descried as ``push forward the drawer handle". The corresponding symmetrical instruction $\tilde{L}$ is obtained through a two-step process: extracting the verb phrase from $L$ and replacing it with an antonym verb phrase, resulting in ``pull backward the drawer handle". The constituency-parser\cite{joshi2018extending} is employed to extract a constituency-based parse tree from $L$, representing its syntactic structure. As an example, the parse tree extraction result of the drawer-close task is shown in Figure~\ref{fig.parse}, where the phrase ``push forward" is identified as the verb phrase (VP). Subsequently, the antonym of the verb phrase is generated utilizing WordNet\cite{miller1995wordnet}, which is a lexical database in English linking words into semantic relations, including synonyms and antonyms. WordNet provides a list of antonyms for each word in the verb phrase ``push forward" ranked in descending order of probability. The highest probability antonym for each word is selected to form the antonym verb phrase ``pull backward". Finally, the symmetrical language instruction $\tilde{L}$ can be generated be substituting the verb phrase in $L$ with the antonym verb phrase.}

\subsection{AIRL Module}

\begin{figure}[t!]
  \centering
  \begin{minipage}[b]{0.49\linewidth}
    \centerline{\quad \textcolor{blue}{\footnotesize Faucet-open reward function}}
    \centerline{\includegraphics[width=\linewidth]{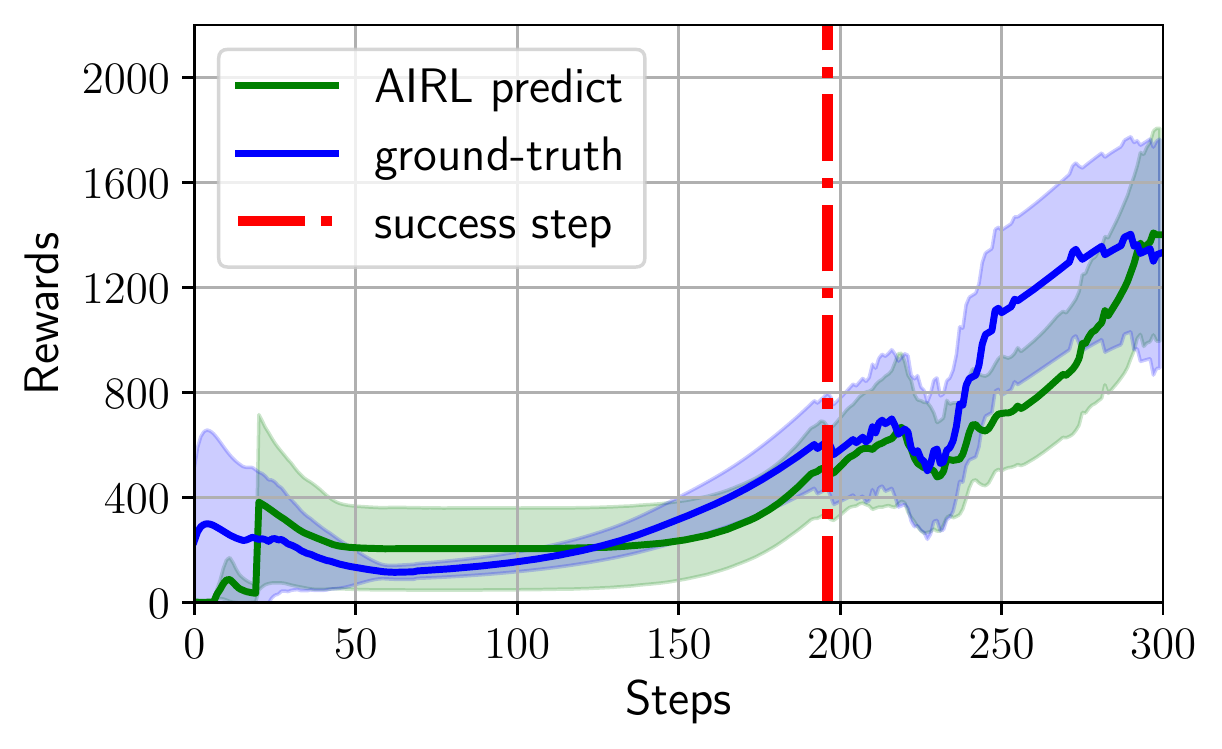}}
  \end{minipage}
  \begin{minipage}[b]{0.49\linewidth}
    \centerline{\qquad \textcolor{blue}{\footnotesize Window-close reward function}}
    \centerline{\includegraphics[width=\linewidth]{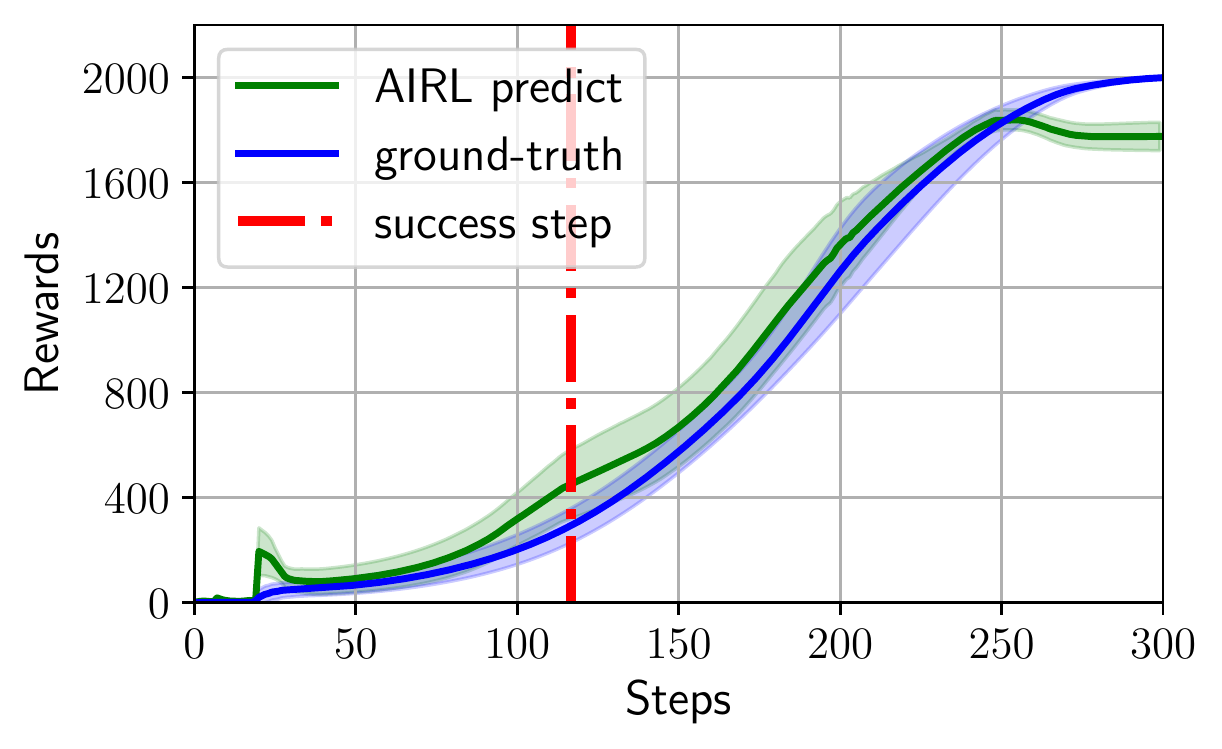}}
  \end{minipage}
  \vspace{-8pt}
  \caption{The performance comparison between the symmetrical task's recovered reward function $\tilde{R}$ and the ground-truth reward function provided by Meta-World\cite{yu2020meta}. The results of each task are generated from the agent by following the optimal AIRL policy in 200 episodes, each of which has 300 time steps. The \textbf{success step} indicates the average time step which the agent successfully solves the task. $\tilde{R}$ shows asymptotic performance with the corresponding ground-truth reward function, indicating that the reward function recovered by $AIRL$ performs as well as the ground-truth reward function. Other comparison results can be found on our website.}\label{fig.reward_airl}
\end{figure}
\footnotetext[1]{\href{https://tumi6robot.wixsite.com/symmetry/}{\textcolor{black}{https://tumi6robot.wixsite.com/symmetry/}}}

\textcolor{black}{The symmetrical trajectory $\tilde{\tau}$ generated by the symmetrical data generator serves two purposes. Firstly, it is stored in \textbf{Buffer} for offline training of an Adversarial Inverse Reinforcement Learning ($AIRL$) network to recover a stable state-only reward function $\tilde{R}$\cite{fu2017learning}. 
Secondly, it functions as an expert demonstration to facilitate the agent to effectively solve $\tilde{T}$ in the symmetrical MDP module\cite{zhou2019watch}.}

\textcolor{black}{In the symmetrical environment with $\tilde{R}$, the agent can effectively associate its behaviors and the language instructions of the symmetrical task through direct interaction from scratch. However, utilizing symmetrical trajectories, derived from the successful trajectories of the original task, for direct agent training may result in sub-optimal performance. This is attribute to the coupling of these symmetrical trajectories with the interaction information during the successful resolution of the original task, and the absence of trial-and-error process, leading to an inaccurate matching between language instructions and behaviors by the agent.}

\textcolor{black}{Prior to offline training of the AIRL network, the algorithm of the agent, MILLION\cite{Million}, is initially trained with all meta-training tasks as depicted in Figure~\ref{fig.task}. During the convergence of MILLION, the successful trajectories of the meta-training tasks are fed to the symmetrical trajectory generator to generate symmetrical trajectories, which are then stored in \textbf{Buffer} until the number of the trajectories exceeds 4000. Subsequently, the AIRL network is trained offline to recover a  state-only reward function for the symmetrical task $\Tilde{T}$. During this process, the SMDP module is inactive. After obtaining the recovered reward function $\Tilde{R}$ for $\Tilde{T}$, the training as outlined in Figure~\ref{fig.framework} is restarted and the SMDP module is activated , while the AIRL module is frozen. The maximum reward of the recovered reward function $\tilde{R}$ is two orders of magnitude smaller than that of the ground-truth reward function provided by Meta-World benchmark\cite{yu2020meta}. To alleviate the impact of the small numerical scale of $\tilde{R}$ in training the agent, the rewards of $\tilde{R}$ are rescaled to (0, 2000). Consequently, the performance of the recovered reward function $\tilde{R}$ was found to be comparable to that of the ground-truth reward function, as shown in  Figure~\ref{fig.reward_airl}.}

\textcolor{black}{The process of grounding the agent's behaviors and language instructions in accordance with that of MILLION \cite{Million}, that is, trial-and-error process, which is time-consuming. To accelerate the trial-and-error process, the SMDP module receives the symmetry trajectory $\tilde{\tau}$ as an expert demonstration. To be specific, the demonstration is a trajectory $D=\{(s_t, a_t)\}$ of $N$ states-action tuples, while in the SMDP module, an entire trajectory of a trail is $\tilde{\tau}=\{(s_t, a_t, \tilde{R}(s_t))\}$ that contains reward function. In our setting, the episode of a MDP has 3 trails, where the first trial involves a demonstration provided to the agent, and the resulting trajectory can be expressed as $\tilde{\tau}=\{(D_t, \tilde{R}(s_t))\}$. The last two trials involve the agent's interaction with the symmetrical environment. In contract to the usage of the demonstration in \cite{zhou2019watch}, the reward function used in the symmetrical environment is recovered by AIRL, rather than from the environment. What's more, the reward function $\tilde{R}$ is state-only, while the reward function in \cite{zhou2019watch} is related to state and action pair, which will leads to the agent's incapacity to adapt to dynamic changes in the environment\cite{fu2017learning}. }

\section{Experiments}

\begin{figure*}[t!]
\footnotesize
    \begin{tikzpicture}[font=\small]
      \matrix (magic) [matrix of nodes]
        {
          Group A:\textcolor{blue}{\footnotesize Meta-training: Reach left} & \textcolor{blue}{\quad\quad \footnotesize Meta-test: Reach right} & \textcolor{blue}{\quad\quad \footnotesize Meta-Training: Push right} &  \textcolor{blue}{\quad\quad \footnotesize Meta-Test: Push left}\\
          \includegraphics[width=0.24\linewidth]{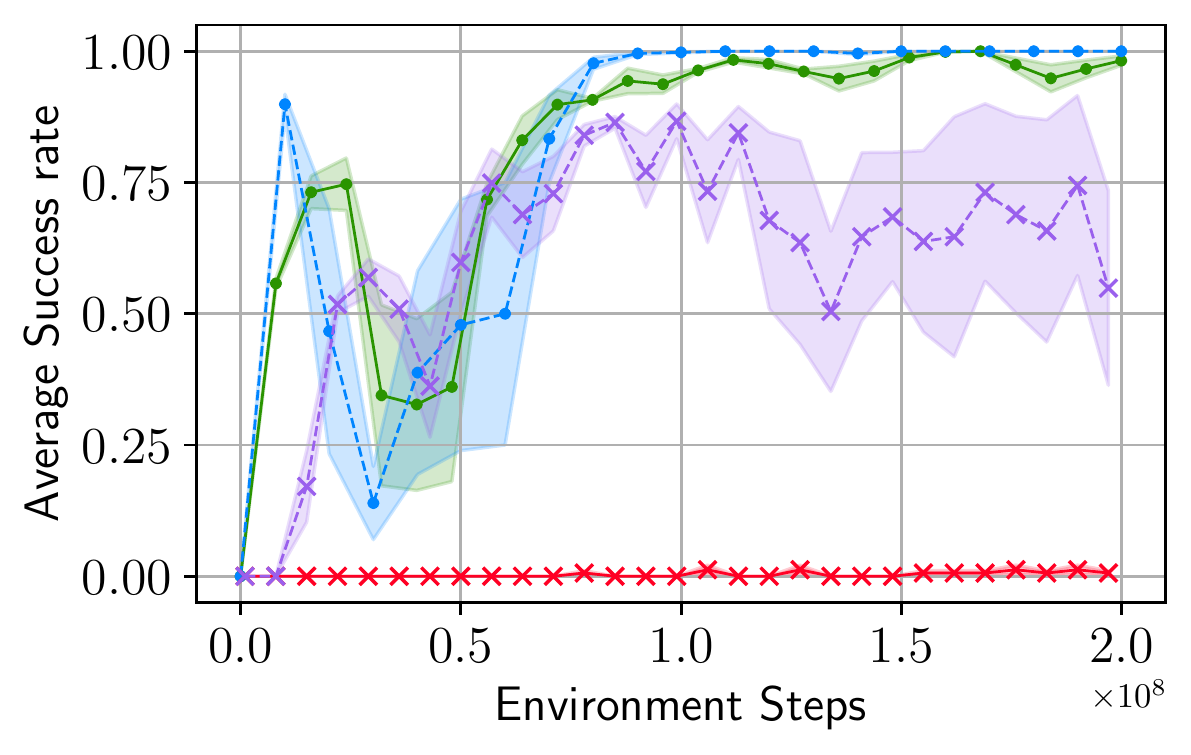} &
          \includegraphics[width=0.24\linewidth]{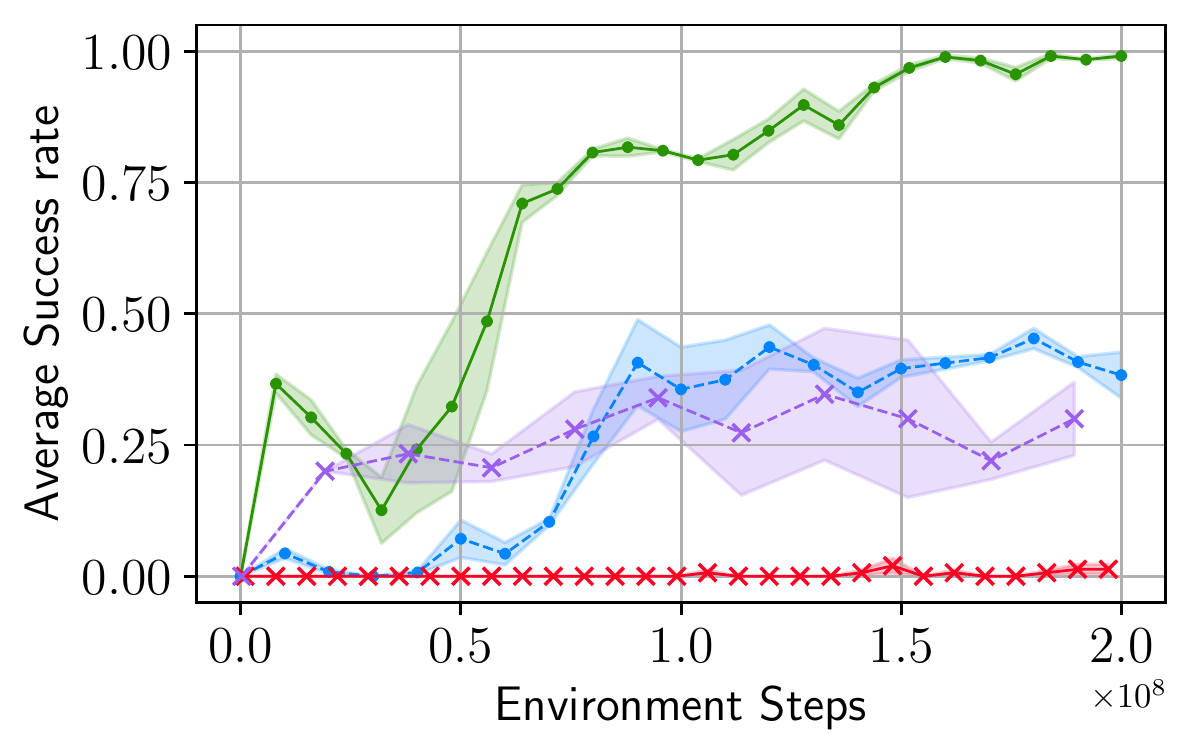} &
          \includegraphics[width=0.24\linewidth]{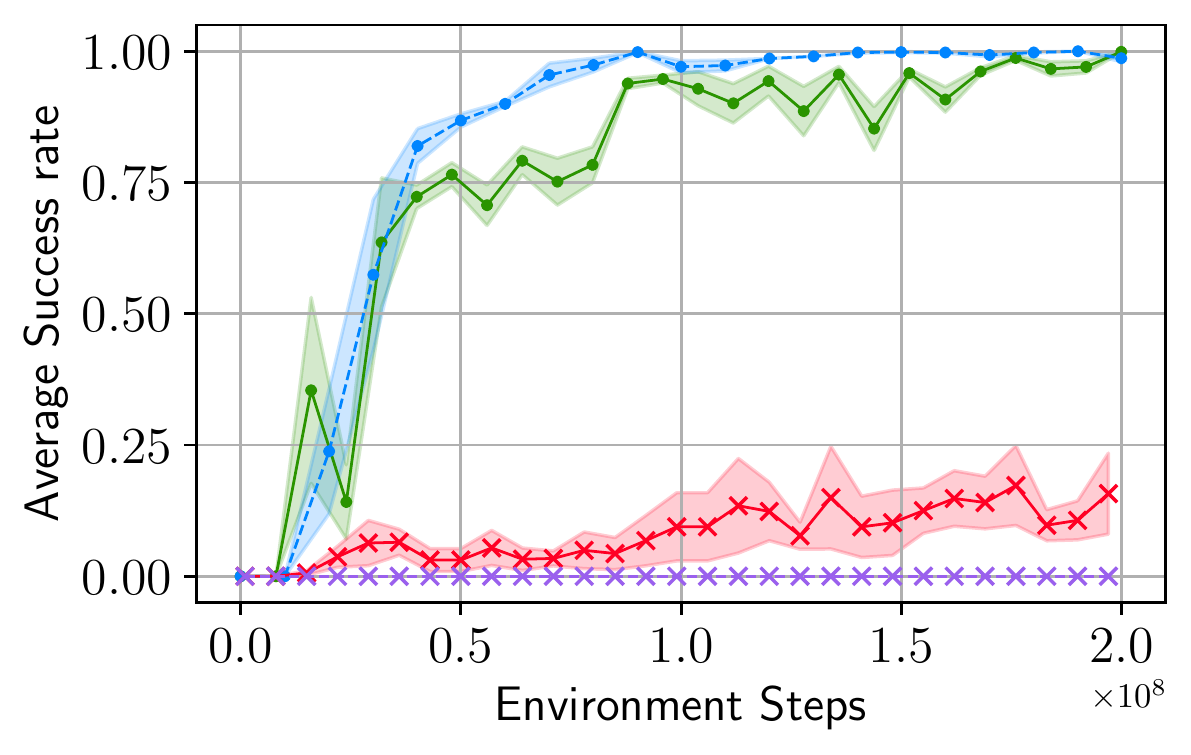} &
          \includegraphics[width=0.24\linewidth]{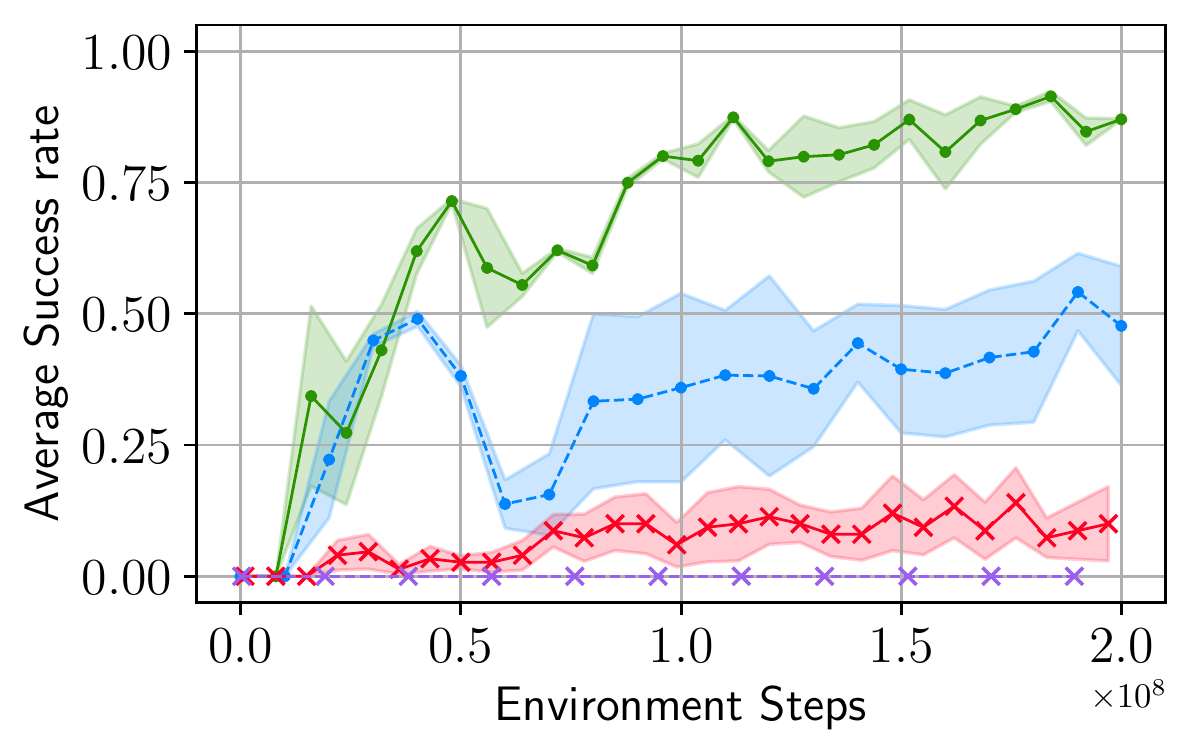} \\
          Group B:\textcolor{blue}{\footnotesize Meta-training: Door open} & \textcolor{blue}{\quad\quad\footnotesize Meta-test: Door close} & \textcolor{blue}{\quad\quad\footnotesize Meta-training: Drawer close} & \textcolor{blue}{\quad\quad\footnotesize Meta-test: Drawer open}\\
          \includegraphics[width=0.24\linewidth]{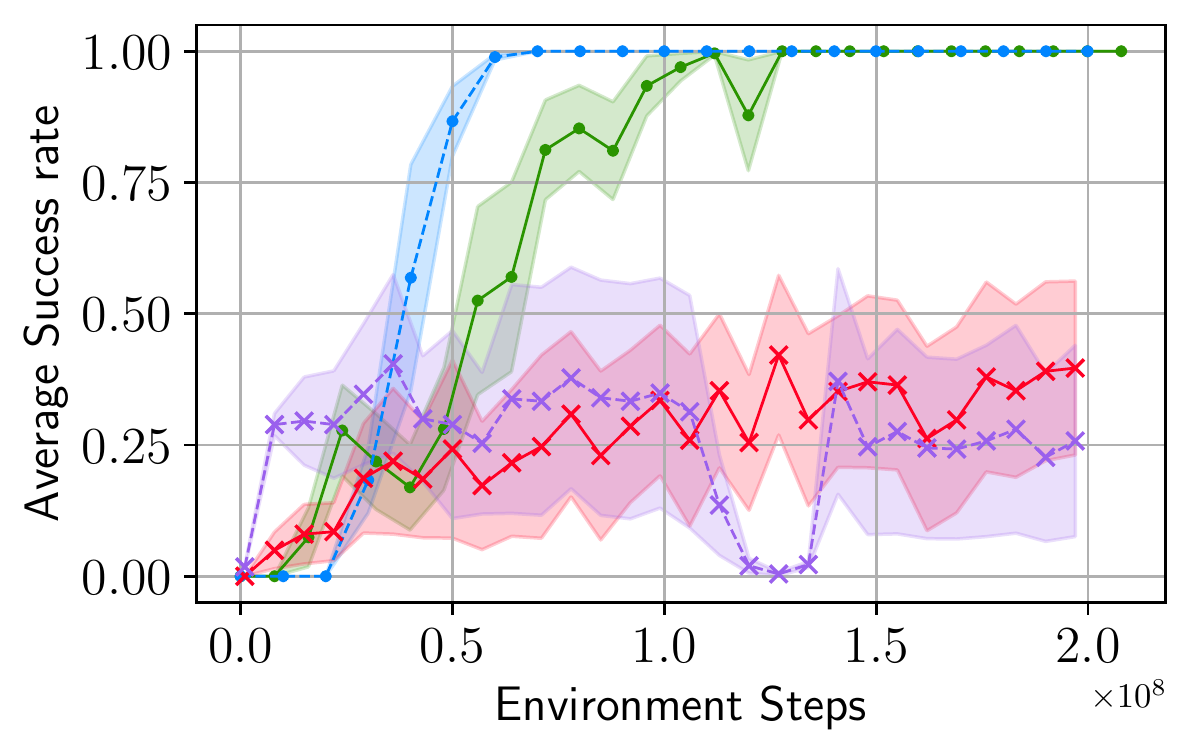} &
          \includegraphics[width=0.24\linewidth]{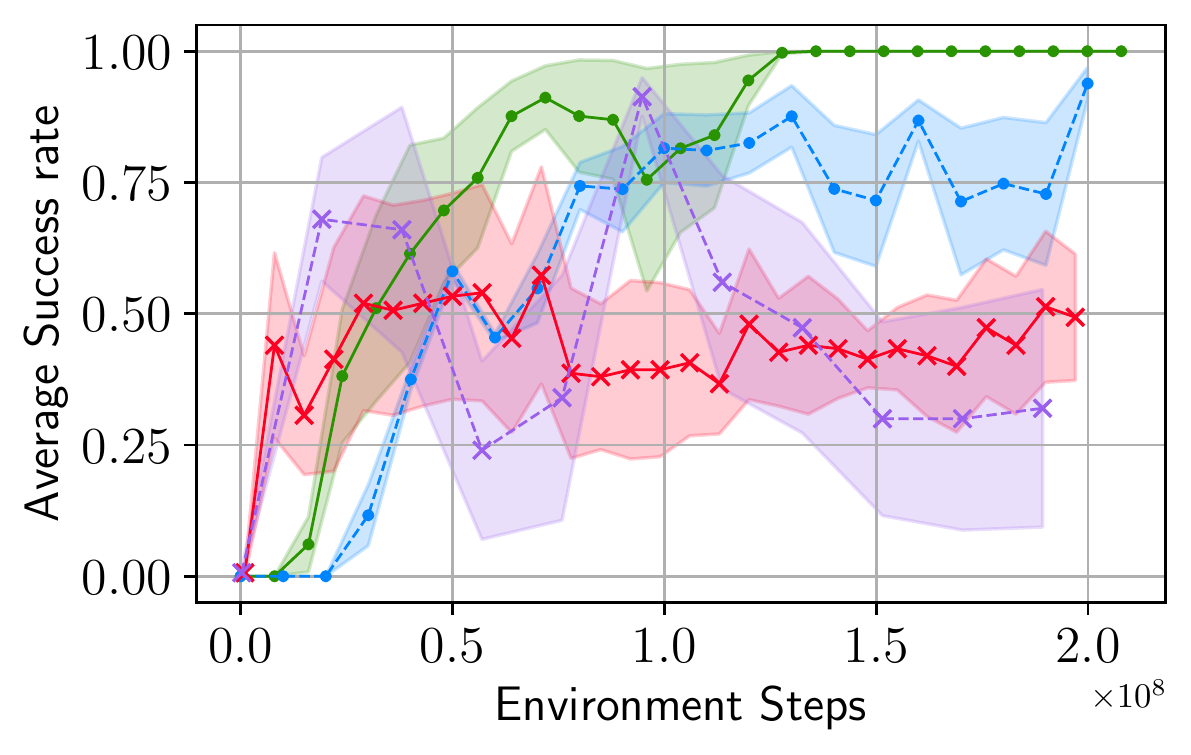} &
          \includegraphics[width=0.24\linewidth]{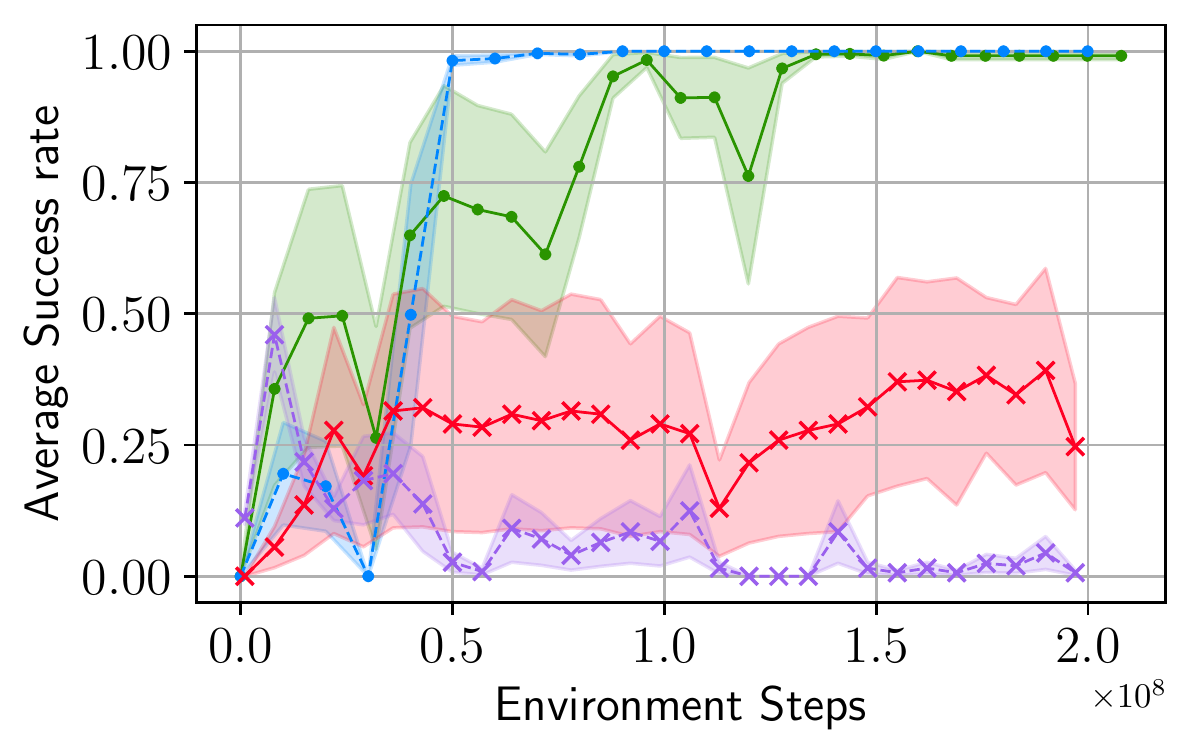} &
          \includegraphics[width=0.24\linewidth]{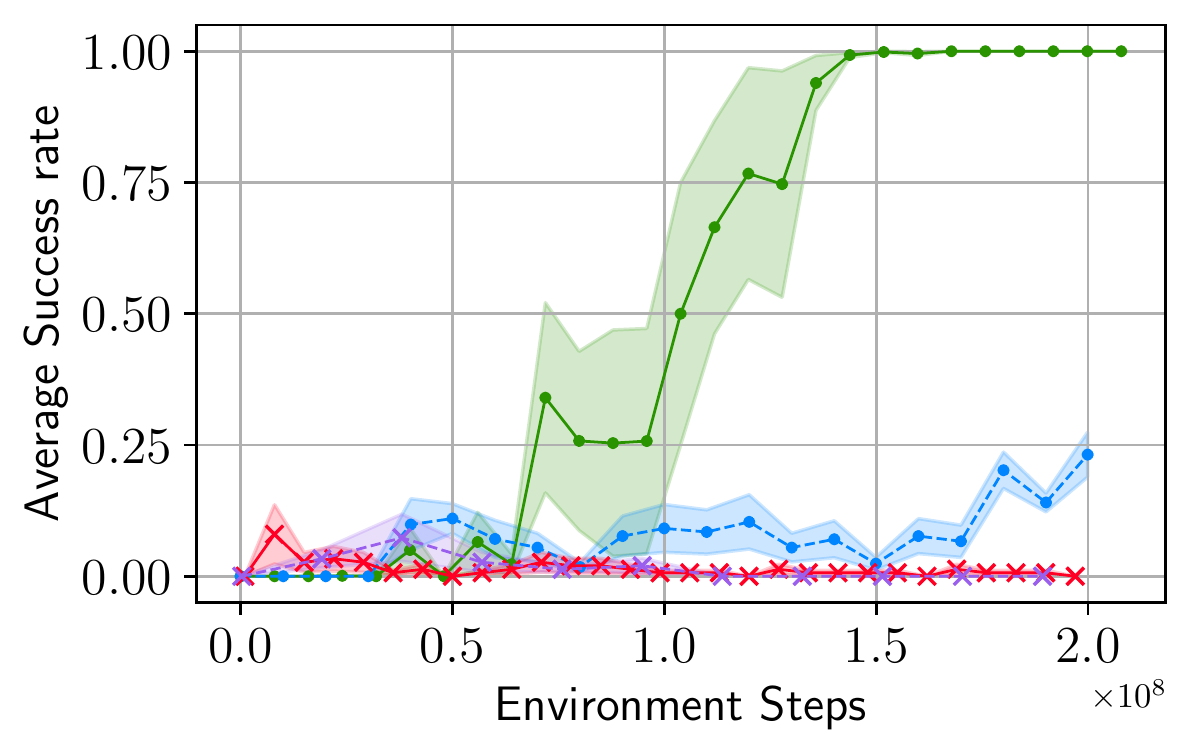} \\
          Group C:\textcolor{blue}{\footnotesize Meta-training: Faucet close} & \textcolor{blue}{\quad\quad\footnotesize Meta-test: Faucet open} & \textcolor{blue}{\quad\quad\footnotesize Meta-training: Window open} & \textcolor{blue}{\quad\quad\footnotesize Meta-test: Window close}\\
          \includegraphics[width=0.24\linewidth]{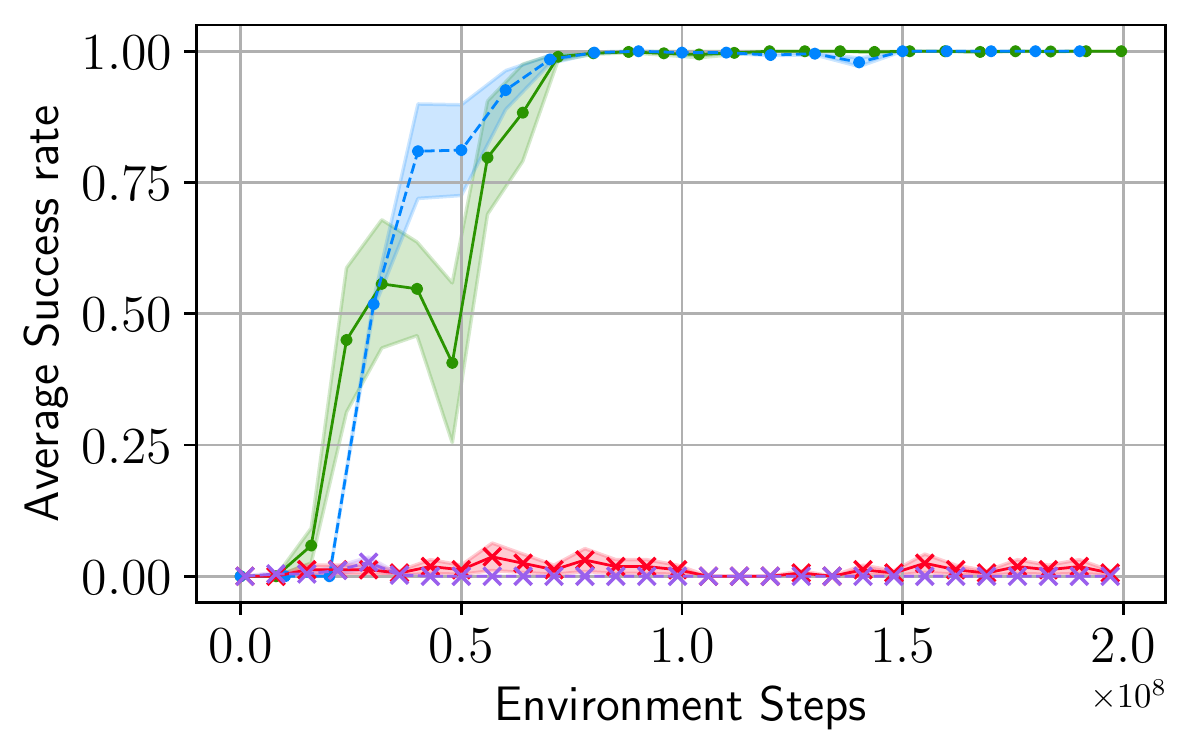} &
          \includegraphics[width=0.24\linewidth]{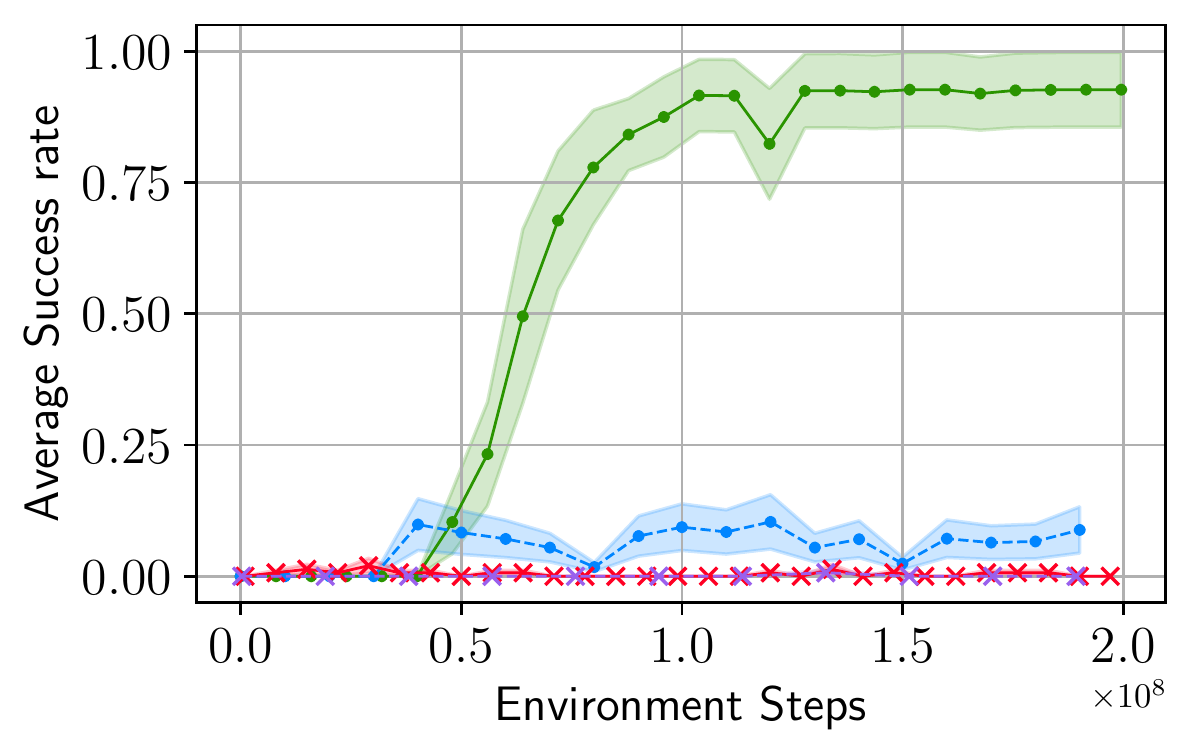} &
          \includegraphics[width=0.24\linewidth]{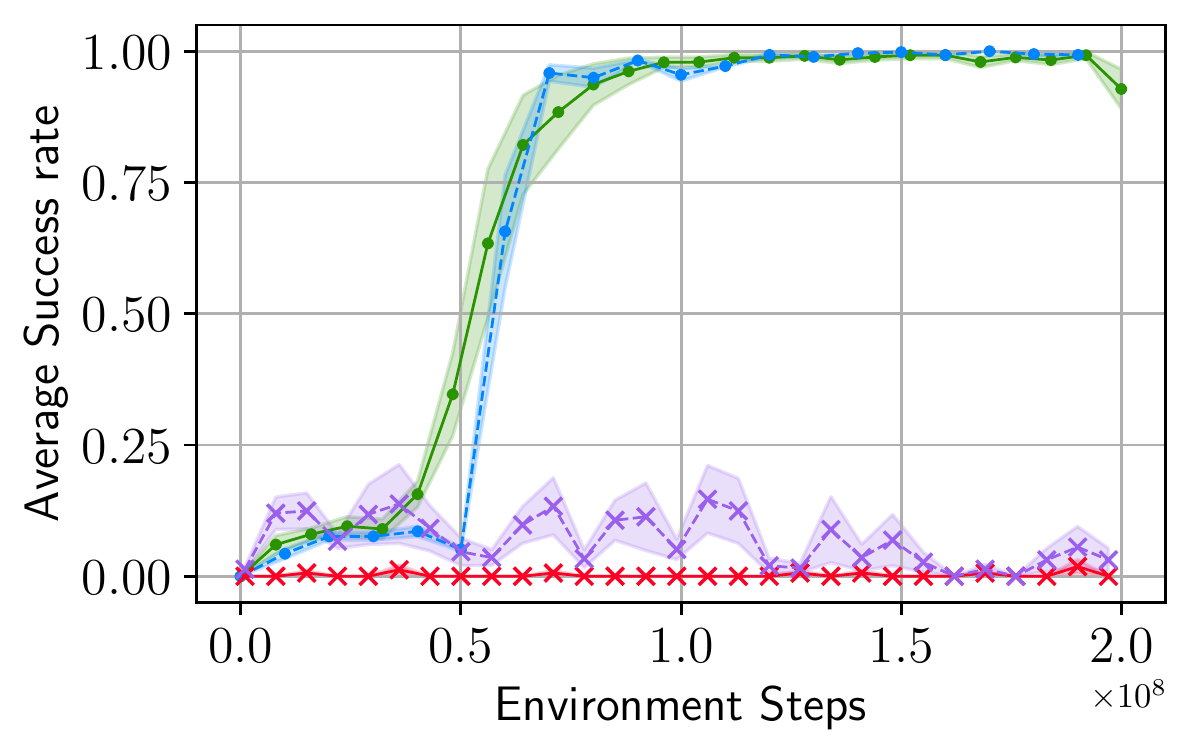} &
          \includegraphics[width=0.24\linewidth]{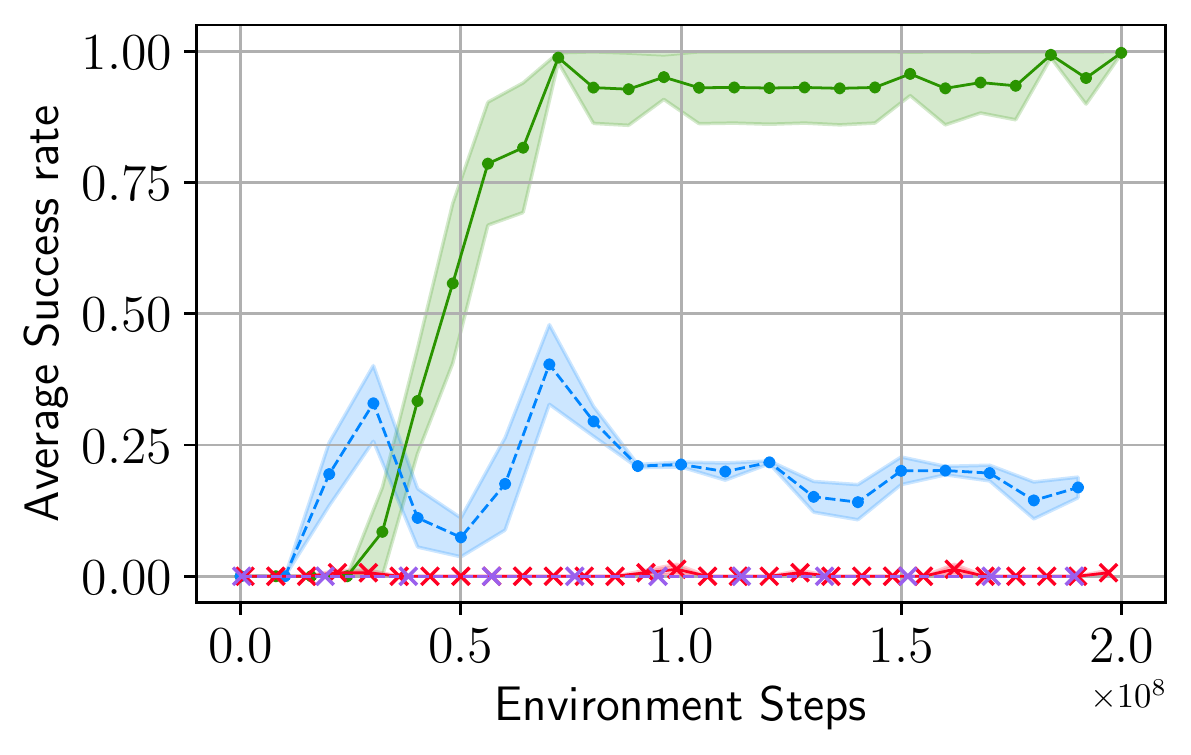} \\
        };
      \matrix[below = 0.01 of magic] (legend) [matrix of nodes]
        {
          \draw[-{Circle[inset=0cm 4]},green, thick](0.0,0)--(0.5,0)
            node[label=right:\footnotesize \textcolor{black}{Ours}]{};&
          \draw[-{Circle[inset=0cm 4]},cyan, dashed, thick](0.0,0)--(0.5,0)
            node[label=right:\footnotesize \textcolor{black}{MILLION}]{}; &
          \draw[-{Rays[inset=0cm]}, red, thick](0.0,0)--(0.5,0)
            node[label=right:\footnotesize \textcolor{black}{MAML-TRPO}]{}; &
          \draw[-{Rays[inset=0cm]}, violet, dashed, thick](0.0,0)--(0.5,0)
            node[label=right:\footnotesize \textcolor{black}{RL$^2$-PPO}]{};\\
        };
        \vspace{-6pt}
    \end{tikzpicture}
    \vspace{-6pt}
\caption{The average meta-training and meta-test result of each task family on 5 seeds. A training group consists of two task families.}
\label{fig.single_task}
\end{figure*}

 \textcolor{black}{In this section, we perform a comparative analysis of our proposed method with three state-of-the-art meta-RL approaches, namely, MILLION\cite{Million}, RL$^2$-PPO and MAML-TRPO. These methods have demonstrated impressive results in Meta-World benchmark\cite{yu2020meta}. Additionally, MILLION is our baseline, which is a language-conditioned meta-RL method and includes only Module 1 (Figure~\ref{fig.framework}). This allows us to perform an ablation study of our method.}

 \textcolor{black}{Our experiments were conducted in two settings. In the first setting, we compared the performance of the methods in three training groups, each of which consists of two task families: Group A for the Reach and Push task family, Group B for the Door and Drawer task family, and Group C for the Faucet and Window task family, as shown in Figure~\ref{fig.single_task}. The training group was designed to investigate the transferability of the methods across a small number of task families, allowing for the potential use of meta-training behaviors to solve meta-test tasks in other task families. For example, the behavior learned from the meta-training task ``open door" could potentially be used to solve the meta-test task ``open drawer". In the second setting, we evaluated the performance of the methods in all task families, where a single policy was used to solve all tasks. The goal of this setting was to investigate the stability of these methods in a multi-manipulation tasks scenario.}


\textcolor{black}{We conducted an evaluation of our proposed method and the baseline, MILLION. While both methods involve the utilization of language instructions to infer meta-test tasks and transfer knowledge obtained from meta-training tasks, our method demonstrated superior performance in both meta-training and meta-test tasks across all training groups compared to MILLION. Conversely, MILLION struggled to successfully complete certain meta-test tasks such as the faucet, drawer, and window task family. This can be attributed to the inability of MILLION's agent to accurately adapt its behavior, meta-learned from meta-training tasks, to the different movement rules of objects in other task families of the same training group. Additionally, this outcome suggests that the abstract attributes of language instruction keywords can impact the agent's generalization capabilities, which corresponds to the problem we mentioned in related work section. In the door task family, although both methods eventually converged to the same success rate in the meta-test, our method required less time and resulted in greater stability. These results highlight the potential of our method to enhance algorithm generalization.}

 \textcolor{black}{Moreover, our method attains comparable success rates as MILLION across all meta-training tasks despite its longer training time. The reason for the slower training time is that the agent's engagement in symmetrical tasks during the meta-training phase, which may decelerate the learning speed of meta-training tasks. Nevertheless, the impact of incorporating the symmetry-aware method (Module 2 and Module 3 shown in Figure~
\ref{fig.framework}) and Symmetrical MDP module is significant, as evidenced by the impressive results attained by the agent across all meta-testing tasks. Specifically, our method outperforms MILLION in tasks such as ``Reach right" and ``Push left" with a success rate 35\% to 60\% higher and tasks such as ``Drawer open," ``Faucet open," and ``Window open" with a success rate 75\% to 80\% higher than that of MILLION.}

\textcolor{black}{Next, we compared our method with the two non language-conditioned meta-RL methods, namely MAML-TRPO and RL$^2$-PPO. The two methods show not good performance on three training groups. The reason is that in these manipulation tasks, the reward function of the meta-training task is distinct from that of the meta-test task, which leads to the agent's inability to transfer the knowledge of the meta-training task to the meta-test task by solely relying on the reward singles. The results indicate that language instructions can improve the generalization of meta-RL.}


 \footnotetext[1]{\href{https://github.com/rlworkgroup/garage}{\textcolor{black}{The implementations: https://github.com/rlworkgroup/garage}}}

 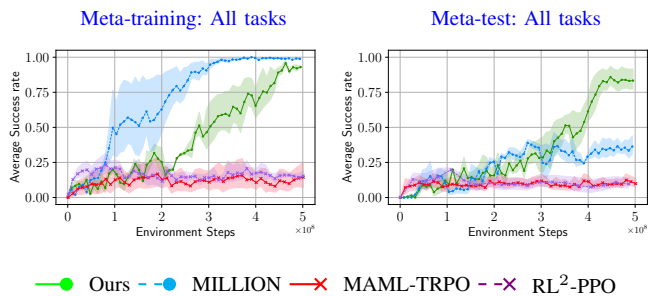
\begin{figure}[htbp!]
\footnotesize
    \begin{tikzpicture}
        \matrix (magic) [matrix of nodes]
        {
          \textcolor{blue}{\quad\quad \footnotesize Meta-training: All tasks} & \textcolor{blue}{\quad\quad \footnotesize Meta-test: All tasks}\\
          \includegraphics[width=0.49\linewidth]{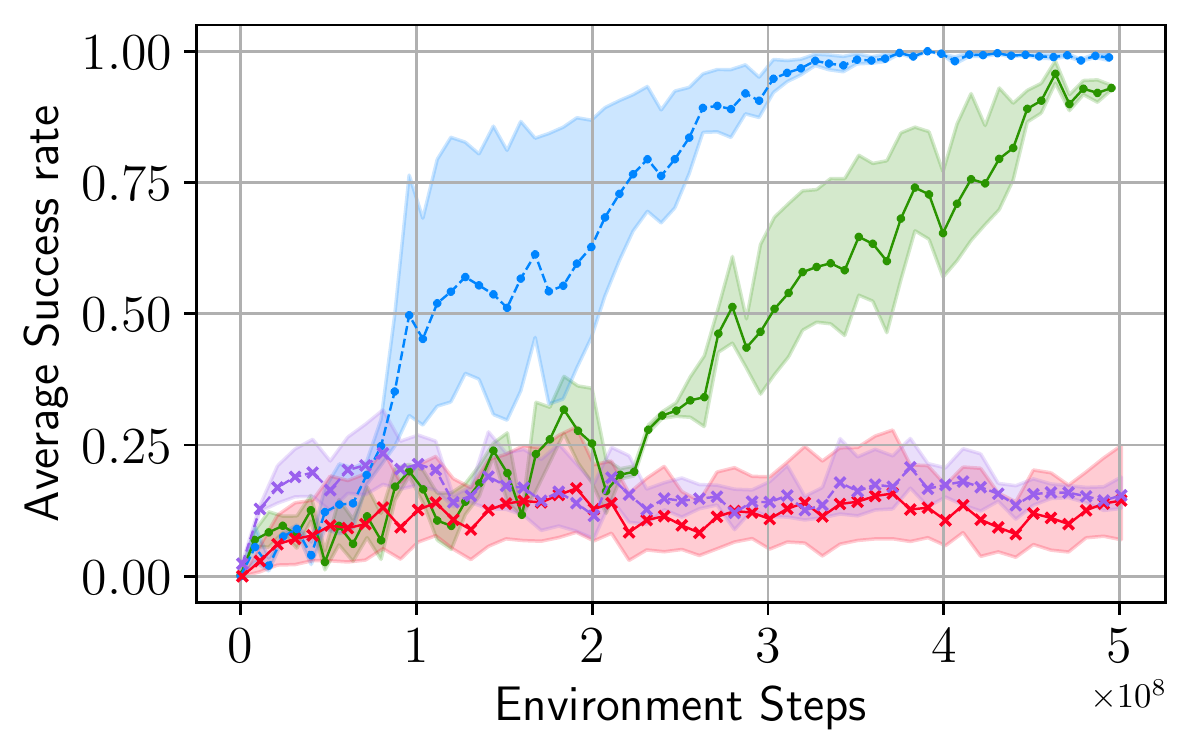} &
          \includegraphics[width=0.49\linewidth]{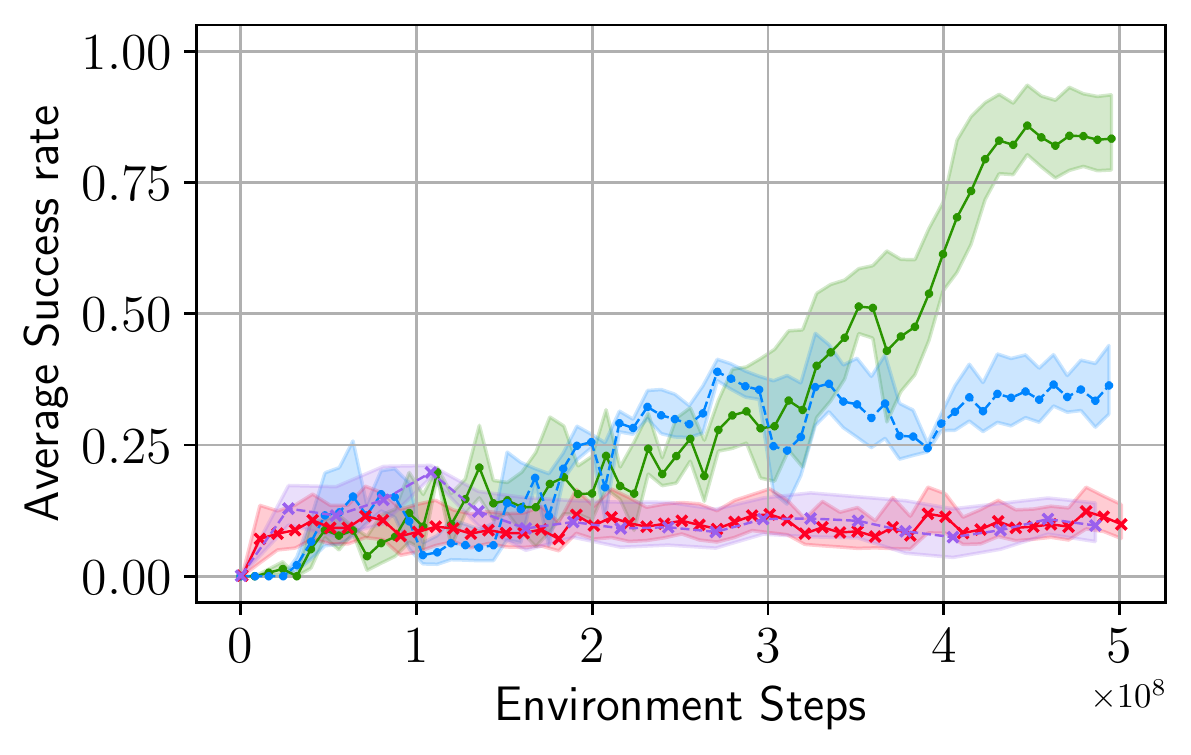}\\
        };
      \matrix[below = 0.01 of magic] (legend) [matrix of nodes]
        {
          \draw[-{Circle[inset=0cm 4]},green, thick](0.0,0)--(0.5,0)
            node[label=right:\footnotesize \textcolor{black}{Ours}]{};&
          \draw[-{Circle[inset=0cm 4]},cyan, dashed, thick](0.0,0)--(0.5,0)
            node[label=right:\footnotesize \textcolor{black}{MILLION}]{}; &
          \draw[-{Rays[inset=0cm]}, red, thick](0.0,0)--(0.5,0)
            node[label=right:\footnotesize \textcolor{black}{MAML-TRPO}]{}; &
          \draw[-{Rays[inset=0cm]}, violet, dashed, thick](0.0,0)--(0.5,0)
            node[label=right:\footnotesize \textcolor{black}{RL$^2$-PPO}]{};\\
        };
        \vspace{-8pt}
    \end{tikzpicture}
    \caption{The average training and test result of the agent solves all task families on 5 seeds. }\label{fig.all_tasks}

\end{figure}

\begin{table}[t!]\normalsize
  \centering
    \caption{Comparison results in all task families.}
    \begin{tabular}{c|c|c} \hline
        Method     & Meta-training & Meta-test    \\
         & (mean$\pm$stderr) & (mean$\pm$stderr) \\\hline
        MILLION\cite{Million}    & \textbf{0.988$\pm$0.006}     & 0.386$\pm$0.06  \\
       RL$^2$-PPO\protect\footnotemark[1]  & 0.154$\pm$0.025  & 0.097$\pm$0.031  \\
       MAML-TRPO\protect\footnotemark[1] & 0.144$\pm$0.075 & 0.099$\pm$0.027 \\
       Our method  & 0.926$\pm$0.037    & \textbf{0.833$\pm$0.059}    \\ \hline
    \end{tabular}
  \label{table:com}
    \vspace{-5pt}
\end{table}

\begin{figure}[t!]
	\centering
	
	\begin{minipage}{0.85\linewidth}
	        \vspace{1pt}
        \centerline{\quad\textcolor{blue}{\footnotesize Language instructions:``Reach to left'', ``Reach to right''}}
        \begin{tikzpicture}
          \centering
          \node[above right, inner sep=0] (image) at (0,0){\includegraphics[width=\linewidth]{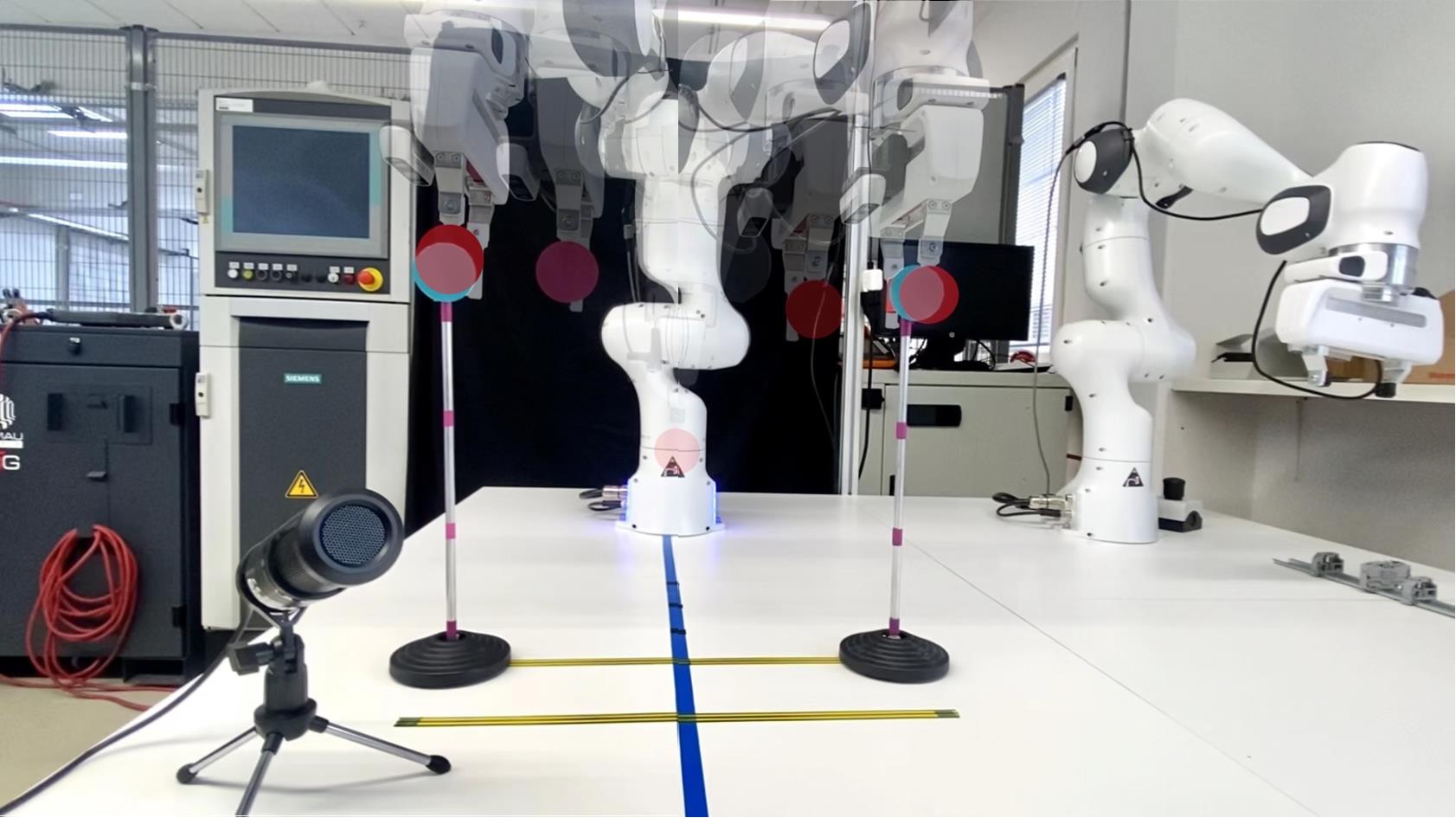}};
          \draw[-stealth, line width=0.3mm, yellow](3.4, 1.95) .. controls(3,2.6).. (2.3,2.7);
          \draw[-stealth, line width=0.4mm, yellow](3.45, 1.95) parabola[bend at end] (4.6,2.6);
          \node(goal2) at (1.45,2.0)[green!70, fill=white] {\footnotesize {right goal1}};
          \draw[arrow1, green!70, line width=0.4mm] (1.9,2.25) -- (2.2,2.6);
          \node(goal1) at (5.3,2)[green!70, fill=white] {\footnotesize {left goal1}};
          \draw[arrow1, green!70, line width=0.4mm] (5,2.25) -- (4.65,2.5);

          \node(start2) at (3.43,1.1)[red,fill=white] {\footnotesize {start}};
          \draw[arrow1, red, line width=0.4mm] (3.43,1.3) -- (3.43,1.7);
          \node(sym) at (4.9,0.25)[blue] {\footnotesize {symmetry line}};
          \draw[arrow1, blue, line width=0.4mm] (4,0.25) -- (3.55,0.25);
          \node(micro) at (0.9,0.35)[blue,fill=white!30] {\footnotesize {microphone}};
          \draw[arrow1, blue, line width=0.4mm] (0.8,0.6) -- (1.2,1.15);
        \end{tikzpicture}	
	\end{minipage}
	




    \caption{Real robot experiments in the reach task family, where the multiple task goals are specified by language instructions through the microphone. In each test case, the goals of the reach-left task and the reach-right task are set to be symmetrical with respect to the symmetry line. }\label{fig.read_exp}
    \vspace{-12pt}
\end{figure}
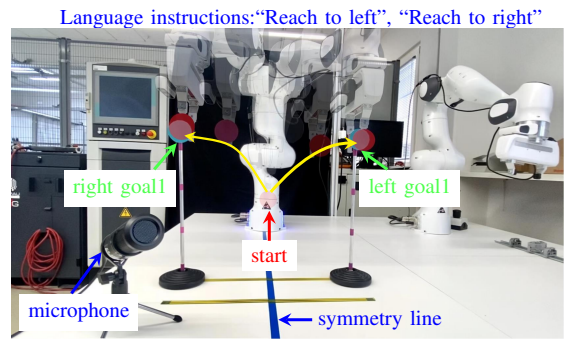

In addition, we also evaluate the four methods in a multi-task setting with all task families, shown as Figure~\ref{fig.all_tasks}. Our method shows asymptotic performance with MILLION in the meta-training tasks, while the average success rate of our method and MILLION in the meta-test tasks are 0.833 and 0.386, respectively, which shows that our method still has stable performance in the multi-tasks scenario. Although our method performs slightly worse than MILLION on the meta-training set, the performance of our method is far better than that of the other three methods on the meta-test set.

Finally, real robot experiments in the reach task family with multiple task goals are carried out, shown as Figure~\ref{fig.read_exp}, where the microphone is used to collect our instructions to direct the agent solve the task. We transfer the policy learned from the Mujoco simulation to the real world and set different task goals to verify the generalization of the policy. The result shows that the policy has good generalization in solving the reach family tasks. What's more, the trajectories of the reach-left and reach-right tasks are nearly symmetrical.

\section{CONCLUSIONS AND OUTLOOK}

In this paper, we propose a dual-MDP meta-RL method combining the symmetry and language instructions to improve the algorithm's generalization and learning efficiency. The agent learns the meta-training task and the symmetrical task in the meta-training stage at the same time, where the symmetrical task is generated by the original task with symmetry and is similar to the meta-test task. Thus the agent can adapt quickly to the meta-test task in the meta-test stage. Learning from symmetry and language is one of the most important forms of human learning, which might bring more inspiration to improve meta-RL.

\normalem
\bibliographystyle{./IEEEtran} 
\bibliography{./main}

\end{document}